\documentclass[sigconf]{acmart}
\renewcommand\footnotetextcopyrightpermission[1]{}
\AtBeginDocument{%
  }

\setcopyright{acmlicensed}
\copyrightyear{2025}
\acmYear{2025}
\acmDOI{XXXXXXX.XXXXXXX}
\acmConference[ACM Conference]{Make sure to enter the correct conference title from your rights confirmation email}{2026}{}

\usepackage{subcaption}
\usepackage{cleveref}
\usepackage{array}
\usepackage{multirow}
\usepackage{stfloats}
\usepackage{enumitem}
\newcolumntype{C}[1]{>{\centering\arraybackslash}p{#1}}

\usepackage[table,xcdraw]{xcolor}

\crefname{figure}{Fig.}{Figs.}
\crefname{table}{Tab.}{Tabs.}
\crefname{section}{Sec.}{Secs.}
\crefname{subsection}{Sec.}{Secs.}
\crefname{equation}{Eq.}{Eqs.}
\crefname{appendix}{Appendix}{Appendices}

\begin{document}

\title[Robotic Manipulation is Vision-to-Geometry Mapping]{Robotic Manipulation is Vision-to-Geometry Mapping: Vision-Geometry Backbones over Language and Video Models}

\author{
  Zijian Song\textsuperscript{1},
  Qichang Li\textsuperscript{1},
  Jiawei Zhou\textsuperscript{1},
  Zhenlong Yuan\textsuperscript{4},
  Tianshui Chen\textsuperscript{3,5},\\
  Liang Lin\textsuperscript{1,2,3},
  Guangrun Wang\textsuperscript{1,2,3,*}
  \\
  \small{
    \textbf{Email:} \{songzj8, liqch33, zhoujw73\} (at) mail2.sysu.edu.cn, yuanzhenlong.yzl (at) alibaba-inc.com, chentianshui (at) gdut.edu.cn,\\ linliang (at) ieee.org, wanggrun (at) gmail.com
  }
}

\affiliation{
    \institution{
        \textsuperscript{1}Sun Yat-sen University; \textsuperscript{2}Guangdong Key Laboratory of Big Data Analysis and Processing;\\
        \textsuperscript{3}X-Era AI Lab; \textsuperscript{4}LongCat, Meituan; \textsuperscript{5}Guangdong University of Technology
    }
    \city{}
    \country{}
}

\renewcommand{\shortauthors}{}

\begin{abstract}
At its core, robotic manipulation is a problem of vision-to-geometry mapping ($f(v) \rightarrow G$). Physical actions—such as reaching, grasping, and orienting—are fundamentally defined by geometric properties like 3D positions, rotations, and spatial relationships. Consequently, we argue that the foundation for generalizable robotic control should be a vision-geometry backbone, rather than the widely adopted vision-language or video models. Conventional Vision-Language-Action (VLA) and video-predictive models rely on backbones pretrained on large-scale 2D image-text or temporal pixel data. While effective, their representations are largely shaped by semantic concepts or 2D priors, which do not intrinsically align with the precise 3D geometric nature required for physical manipulation. Driven by this insight, we propose the Vision-Geometry-Action (VGA) model, which directly conditions action generation on pretrained native 3D representations. Specifically, VGA replaces conventional language or video backbones with a pretrained 3D world model, establishing a seamless vision-to-geometry mapping that translates visual inputs directly into physical actions. To further enhance geometric consistency, we introduce Progressive Volumetric Modulation and jointly train action and 3D property prediction to preserve geometric representations. Extensive experiments validate the effectiveness of our approach. Across simulation benchmarks, VGA outperforms leading VLA, 3D-VLA, and WAM baselines, including $\pi_{0.5}$, OpenVLA-OFT, GeoVLA, and Motus, demonstrating precise spatial manipulation and robust generalization. In real-world deployments, VGA surpasses $\pi_{0.5}$ under unseen viewpoints and accurately follows language instructions for target grasping. These results highlight that operating on native 3D representations—rather than relying primarily on language or video priors—offers a promising direction toward generalizable physical intelligence.
Project page: \url{https://hcplab-sysu.github.io/VisionGeometryActionModel/}.
\end{abstract}




\maketitle


\renewcommand{\floatpagefraction}{0.8}
\section{Introduction}


\begin{figure}[t]
    \centering
    \includegraphics[width=1.0\linewidth]{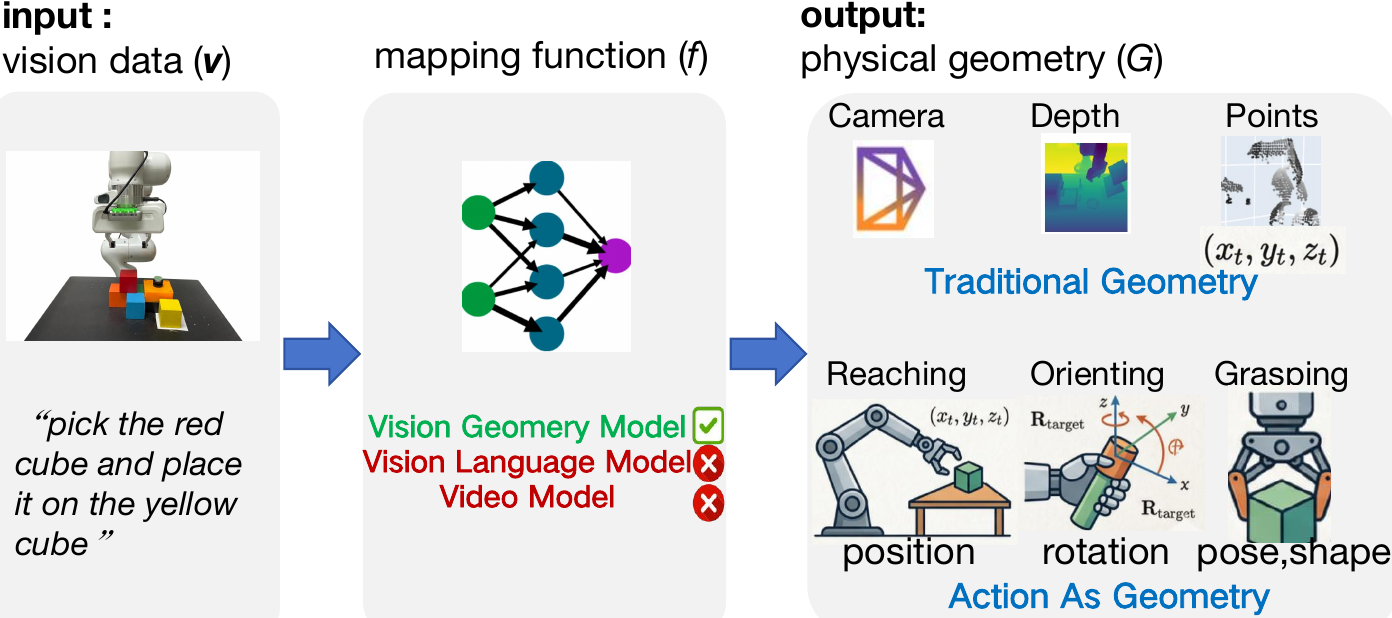}
    \caption{Robotic manipulation as vision-to-geometry mapping ($f(v) \rightarrow G$). Physical actions like reaching, grasping, and orienting are inherently driven by geometric properties, such as 3D position, rotation, and spatial relationships. Therefore, we argue that a vision-geometry backbone provides a stronger foundation for generalizable robotic control than prevalent vision-language or video models.}\label{fig:compare_rep}
\end{figure}

Fundamentally, robotic manipulation is a problem of vision-to-geometry mapping ($f(v) \rightarrow G$). Physical actions—such as reaching, grasping, and orienting—are inherently defined by precise 3D geometric properties and spatial relationships (see Fig. \ref{fig:compare_rep}). However, the recent pursuit of generalizable robotic control has been heavily dominated by Vision-Language-Action (VLA) models~\cite{zitkovich2023rt, kim2024openvla, nvidia2025gr00t, wang2025vla, zhan2025mathcal} and video-predictive policies. Driven by progress in generative AI, video-centric approaches such as World Action Models aim to capture physical dynamics by jointly predicting future frames and actions using massive video diffusion backbones~\cite{ye2026world}. Together, these language and video approaches typically rely on backbones pretrained on massive internet-scale 2D image-text or temporal pixel data. While these models excel at interpreting linguistic semantics, anticipating temporal sequences, and generating actions across diverse environments~\cite{black2024pi0, xu2025a0, bu2025agibot, li2025vla, chen2025villa, pi05vision, zhan2026stable}, they are fundamentally optimized for generating semantic concepts or predicting 2D temporal changes, rather than reasoning about spatial reality\let\thefootnote\relax\footnote{*Corresponding author: Guangrun Wang}.

This reliance on language and video models introduces a fundamental discrepancy between the 2D pretraining of current backbones and the 3D nature of physical manipulation. Manipulation requires genuine spatial intelligence~\cite{chen2024spatialvlm, mao2025spatiallmtraininglargelanguage, wang2025indvissgg, liao2025improved}—the ability to reason over volume, geometry, and physical object relationships. Because the backbones of VLA and video models are shaped by 2D priors, they tend to overfit visual patterns rather than capturing true 3D dynamics~\cite{chen2024sugar, qu2025spatialvla, sun2025geovla, chen2026radar}. While predicting dense temporal changes offers an implicit proxy for physics, the representation remains locked in pixel space. As shown in~\cref{fig:compare_rep}, a 2D representation may interpret object motion primarily as pixel displacement, struggling to distinguish whether an object is being pushed or lifted, or accurately reason about its spatial relations within a scene. This misalignment between learned representations and 3D physical actions ultimately limits robust generalization in complex environments.

An intuitive remedy is to incorporate explicit 3D geometric information; yet, existing adaptations remain suboptimal. One practice introduces 3D inputs, such as depth maps or point clouds~\cite{chen2024sugar, jia2024lift3d, ze20243d, sun2025geovla, yuan2025depthvla, li20253ds, li2026pointvla}. While providing precise geometry, these methods rely on extra sensors that introduce noise, increase fusion complexity, and raise hardware costs. Alternatively, recent efforts prepend 3D-aware encoders to a VLM backbone~\cite{qu2025spatialvla, lin2025evo, vuong2025improving, abouzeid2025geoaware, rao2026augvla}. However, as indicated by recent studies~\cite{liu2025deconstructing, li2025does, liu2026spatial}, the latent representations of VLMs remain stubbornly 2D-centric. Consequently, rich 3D information is projected back into a flat representation space. This creates a flawed \emph{3D-2D-3D} transformation loop, where 3D geometry features are forced through a 2D latent bottleneck before being decoded back into 3D actions.

To break this bottleneck, we aim to build a robotic foundation model strictly aligned with the $f(v) \rightarrow G$ paradigm, where perception, reasoning, and action are physically aligned within a shared native 3D representation space. We propose the Vision-Geometry-Action (VGA) model, which replaces conventional 2D language or video backbones with a pretrained 3D world model, i.e., VGGT~\cite{wang2025vggt}. VGA takes multi-view observations as input and produces native spatial representations, inheriting strong 3D priors from VGGT. By conditioning action prediction entirely on these representations, VGA establishes a seamless vision-to-geometry mapping that translates visual inputs directly into physical actions. To ensure these spatial priors effectively guide action generation, we introduce a Progressive Volumetric Modulation module that bridges the backbone and decoding heads, facilitating a high-fidelity flow of geometric information. Furthermore, inspired by the World-Action Model (WAM) paradigm~\cite{liang2025video, li2025unified, zhang2025dreamvla, ye2026world, li2026causal}, we adopt a joint-training strategy where shared 3D representations predict both actions and 3D properties. During inference, decoupled decoding ensures execution efficiency while retaining deep spatial awareness.

We evaluate VGA through extensive experiments in both simulated and real-world environments.
Across LIBERO, RoboTwin2.0, and LIBERO-Plus, VGA outperforms representative baselines, including $\pi_{0.5}$~\cite{pi05vision}, GeoVLA~\cite{sun2025geovla}, and Motus~\cite{bi2025motus} without relying on VLM backbones or additional 3D sensors, highlighting the efficacy of a vision-geometry backbone for precise manipulation. Moreover, quantitative analysis confirms the high fidelity of VGA's learned 3D properties.
In physical robot deployments, VGA performs strongly in standard setups and language-grounded manipulation tasks, while exhibiting remarkable zero-shot generalization to unseen camera views.
These results demonstrate that native 3D representations can effectively bridge the perception-action gap and support robust manipulation under substantial observational variations.

The major contributions of this paper are summarized as follows:\begin{itemize}[leftmargin=1.5em]
    \item We formalize robotic manipulation as a vision-to-geometry mapping ($f(v) \rightarrow G$) and propose the Vision-Geometry-Action (VGA) model, moving beyond 2D pattern matching toward physically grounded perception and action.
    \item We develop a unified 3D-centric architecture that prioritizes a Vision-Geometry backbone over conventional language or video models, integrating Progressive Volumetric Modulation and joint training to entirely bypass the representation bottleneck of 2D-centric processing.
    \item Extensive experiments across simulation and real-world settings demonstrate VGA’s strong performance and generalization, validating the superiority of 3D representations for robotic manipulation.
\end{itemize}

\section{Related Work}

\begin{figure*}[t]
    \centering
    \includegraphics[width=0.9\linewidth]{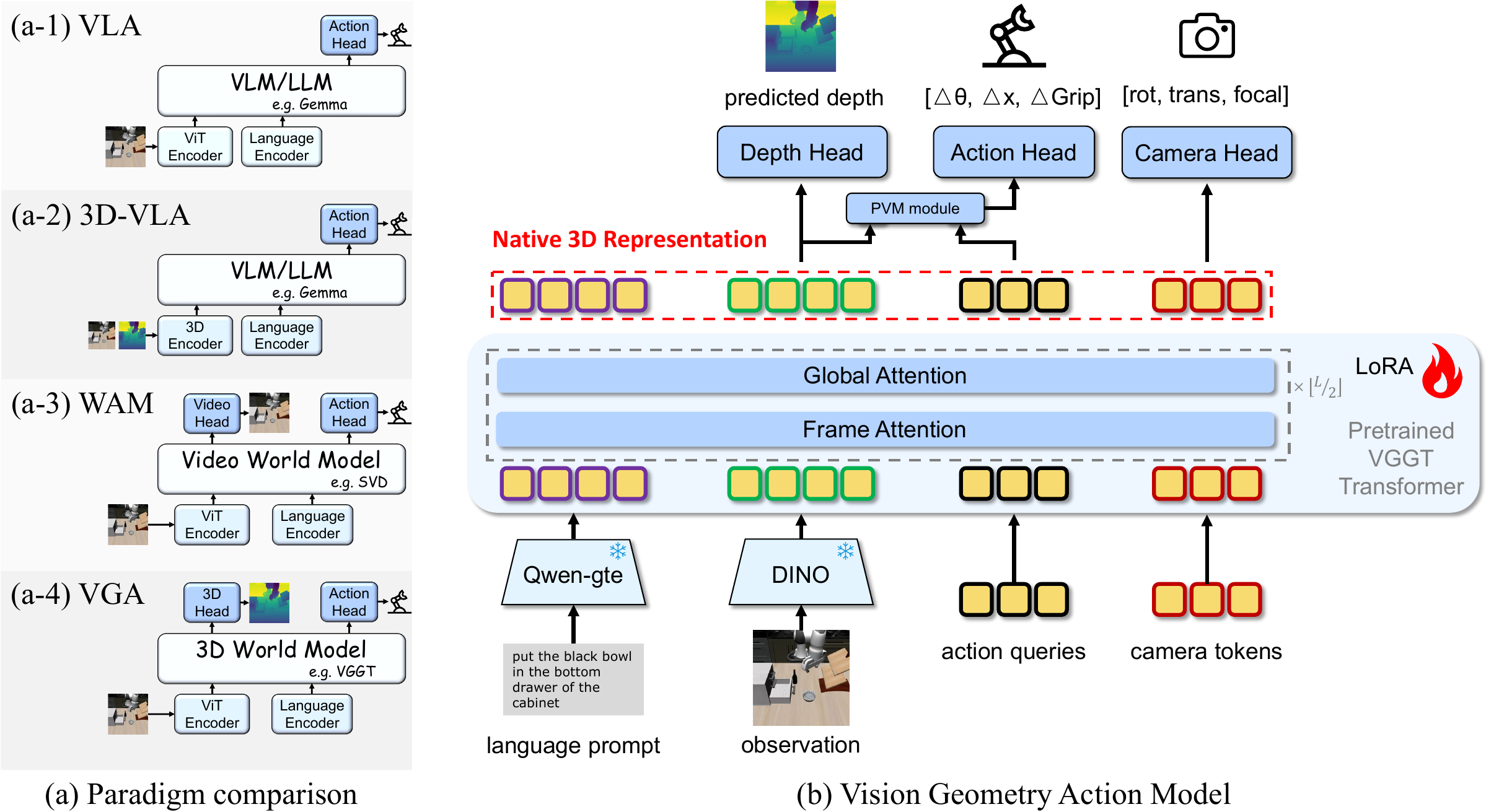}
    \caption{\underline{Overview of our VGA model.}
    (a) The left column compares our VGA framework with representative robot learning paradigms.
    VGA differs from them by leveraging a pretrained 3D world model as the backbone, providing native 3D representations aligned with physical actions.
    (b) The right column illustrates the workflow of the VGA model.
    Multimodal inputs are tokenized into a unified sequence and processed by a pretrained VGGT transformer with alternating attention.
    The resulting latent features are then mapped by task-specific heads to produce multimodal outputs, each with corresponding supervision.
    }
    \label{fig:pipline}
\end{figure*}

This work relates to VLA, 3D-VLA, and WAM.
\cref{fig:pipline}-(a) provides a comprehensive comparison across these paradigms.

\subsection{Vision-Language-Action Models}

Vision-Language-Action (VLA) models have become a dominant paradigm for general robot control by adapting pretrained vision-language models for action generation~\cite{zitkovich2023rt, team2024octo, wen2025tinyvla, wen2025dexvla, liu2025hybridvla, bu2025univla}.
Recent works further explore larger models~\cite{bjorck2025gr00t, xu2025a0, li2025vla, nvidia2025gr00t, zhou2026tag}, diverse data~\cite{kim2024openvla, black2024pi0, bu2025agibot, pi05vision}, and discrete diffusion formulations~\cite{li2025situ, zhan2025mathcal}.
However, under the $f(v) \rightarrow G$ framework, their representations are shaped by the 2D image-text data used in pretraining and lack the precise spatial awareness required for rigorous physical interaction~\cite{chen2024spatialvlm, qu2025spatialvla, zhou2025physvlm, liu2026spatial}. Rather than relying on 2D semantic concepts, we argue that action generation must be directly conditioned on native geometric structures. By replacing language backbones with a vision-geometry foundation, our approach aligns control strictly with spatial reality, leading to significantly improved robustness and generalization.


\subsection{3D Perception with VLA Models}

To mitigate the limited spatial awareness of standard VLA frameworks, existing works incorporate 3D information through 3D position encodings~\cite{qu2025spatialvla}, embeddings from estimated point clouds~\cite{rao2026augvla}, auxiliary geometry encoders~\cite{lin2025evo, vuong2025improving, abouzeid2025geoaware, yang2026abot, liu2026activevla}, or external spatial sensors~\cite{chen2024sugar, jia2024lift3d, ze20243d, sun2025geovla, yuan2025depthvla, li20253ds, li2026pointvla}.
However, these approaches still rely on VLM backbones whose representations are primarily shaped by 2D image-text pretraining. As a result, rich 3D information is inevitably projected into 2D-centric latent spaces, creating a flawed 3D-2D-3D transformation loop.
In stark contrast, VGA adopts a pretrained 3D world model as the backbone, ensuring that representations remain firmly grounded in geometric structure throughout the entire perception-to-action pipeline.


\subsection{World Action Models}

Our work is also closely related to World Action Models (WAMs), also known as Video Action Models (VAMs), which jointly predict future frames and actions~\cite{hu2024video, liang2025video, pai2025mimic, zhu2025unified, li2025unified, zhang2025dreamvla, zhao2025cot, shen2025videovla, kim2026cosmos, song2025physical, song2026learning, ye2026world, li2026causal}.
While WAMs offer a powerful proxy for physics, their backbones remain fundamentally locked in 2D pixel space, optimizing for temporal changes rather than structural 3D reality.
Inspired by the joint-training philosophy of WAMs, we propose to jointly predict actions and 3D geometric properties to encourage geometric-to-action mapping. By integrating a 3D world model rather than a video diffusion model, VGA establishes a unified action-geometry foundation that captures the physical essence of manipulation.

\section{Preliminaries}
\label{sec:preliminaries}

VGGT~\cite{wang2025vggt} is a 3D geometric foundation model built upon a unified feed-forward transformer.
Given multi-view RGB observations, VGGT produces native geometric representations, including camera parameters, depth maps, point maps, and dense correspondence features.
Pretrained on large-scale multi-view 3D datasets such as Co3Dv2\cite{reizenstein21common} and BlendMVS\cite{yao2020blendedmvs}, VGGT develops strong spatial priors that are essential for understanding physical geometry.

The core of the VGGT architecture is a transformer-based backbone that employs an Alternating-Attention mechanism.
This design interleaves frame-wise local attention, which processes spatial details within individual camera views, and cross-frame global attention, which aggregates information across multiple perspectives to build a unified 3D understanding.
On top of the resulting representations, VGGT adopts a set of specialized decoding heads for different geometric predictions.
Detailed formulations and architectural specifics are provided in the Appendix.

In this work, we employ the pretrained VGGT as our foundational backbone, directly inheriting its pretrained weights to leverage the rich spatial intelligence acquired during its pretraining.
We retain the core architectural design of VGGT, enabling it to provide native 3D representations specifically for robotic manipulation.

\section{Method}

\subsection{Overview}

The core philosophy of Vision Geometry Action model (VGA) is to leverage a pretrained 3D world model as the backbone, enabling a seamless vision-to-geometry mapping that translates visual inputs directly into physical actions.
As illustrated in~\cref{fig:pipline}, at each timestep $t$, VGA takes multi-view RGB observations $\left\{I_{t}^{i}\right\}_{i=1}^{N}$, a language instruction $\ell$, and robot proprioception $S_t$ as input, where $N$ denotes the number of views.
These multimodal inputs are processed by the pretrained VGGT backbone to obtain native 3D representations $\mathbf{V}_t\in \mathbb{R}^{T \times D}$, where $T$ is the total number of aggregated tokens and $D$ is the token dimension.
The resulting representations are fed into decoupled decoding heads to jointly predict an action chunk $\mathbf{a}_{t:t+C}$ and auxiliary 3D properties $\left\{\mathbf{g}_{t}^{i}, D_{t}^{i}\right\}_{i=1}^{N}$, where $C$ denotes the action chunk size, and $\mathbf{g}_{t}^{i} \in \mathbb{R}^{9}$ and $D_{t}^{i} \in \mathbb{R}^{H \times W}$ represent the camera parameters and depth map of the corresponding view, respectively.

\subsection{Native 3D Representation}

At each timestep $t$, VGA first converts multi-view observations, language instructions, and proprioception into a unified token sequence.
Specifically, each image view $I_{t}^{i}$ is tokenized by a DINO encoder~\cite{caron2021emerging} into $K$ patch tokens, which are used as visual embeddings.
The proprioception $S_t$ is projected into a $d$-dimensional embedding $e_{prop}$ through an MLP.
The language instruction $\ell$ is first encoded by Qwen-GTE~\cite{li2023towards} and then projected through an MLP before being integrated with other input modalities.
Following VLA paradigms~\cite{zhao2025cot, zhang2025dreamvla}, VGA introduces learnable action queries $q_{act}$ to aggregate manipulation-related information from the multimodal inputs.
In addition, VGA adopts learnable camera tokens $q_{cam}$ from VGGT to capture camera information.
All modality embeddings are then concatenated into a unified token sequence:
\begin{equation}
\tilde{X}^{(l)} = \left[X_1^{(l)}, \dots, X_k^{(l)}, \dots, X_N^{(l)}, X_{\text{lang}}^{(l)}, X_{\text{act}}^{(l)}\right],
\end{equation}
where $X_k^{(l)}$ denotes the input features of the $k$-th modality at layer $l$; $[\cdot, \cdot]$ denotes the concatenation; $k \in \{1, \dots, N, \text{lang}, \text{act}\}$ indexes camera views, language tokens, and action queries.

The unified token sequence is then processed by the VGGT backbone through $\lfloor L/2 \rfloor$ Alternating-Attention stages, alternating between frame-wise local attention at even layers ($l=2m$) and cross-modal global attention at odd layers ($l=2m+1$).
At even layers, local attention is applied independently within each modality:
\begin{equation}
X_k^{(l)} = \text{Attention}(Q=X_k^{(l-1)}, K=X_k^{(l-1)}, V=X_k^{(l-1)}).
\end{equation}
At odd layers, global attention is applied over the entire token sequence $\tilde{X}^{(l-1)}$:
\begin{equation}
\tilde{X}^{(l)} = \text{Attention}(Q = \tilde{X}^{(l-1)}, K = \tilde{X}^{(l-1)}, V = \tilde{X}^{(l-1)}).
\end{equation}

Ultimately, this process yields the unified 3D representations $\mathbf{V}_t = \tilde{X}^{(L)}$ that capture the underlying scene geometry as well as its alignment with task-relevant semantics.

\subsection{Joint Training and Decoupled Inference}

Following the joint-training paradigm of World Action Models (WAMs)~\cite{cen2025worldvla, li2025unified, liang2025video}, VGA jointly decodes the shared 3D representations into actions and geometric properties through task-specific heads.
Specifically, it comprises an action head and two auxiliary heads for camera and depth prediction, providing multi-task supervision for the shared representations.

The action head is an $L_a$-layer regression transformer with $L_a$ layers ($L_a=L$), similar to OpenVLA-OFT~\cite{kim2025fine}.
To enable temporal look-ahead and smoother control, it predicts chunks of $C=8$ actions from learnable noise embeddings $\mathbf{z} \in \mathbb{R}^{C \times D}$.
Conditioned on $\mathbf{V}_t$ through PVM, these embeddings are processed by $L_a$ transformer blocks and projected by a linear layer $\phi$ into $\hat{\mathbf{a}}_{t:t+C} \in \mathbb{R}^{C \times A}$, where $A$ is the action dimension:
\begin{equation}
\hat{\mathbf{a}}_{t:t+C} = \phi(\mathbf{z}^{(L_{a})}).
\end{equation}

The auxiliary camera and depth heads follow the original VGGT design and provide explicit geometric supervision. The camera head regresses the parameters $\mathbf{g}_i$ from the evolved camera token $\mathbf{t}_{i,\mathbf{g}}^{(L)}$.
The depth head leverages a Dense Prediction Transformer (DPT) module to reconstruct the pixel-wise depth map $D_i$ by hierarchically aggregating multi-scale backbone tokens $\mathbf{t}_i^{(s)}$ through a reassembly mapping $\phi_s$ and a fusion operator $\Psi$.
They can be written as:
\begin{equation}
\mathbf{g}_i = [\mathbf{r}_i, \mathbf{t}_i, \mathbf{f}_i] = \text{MLP}(\mathbf{t}_{i,\mathrm{g}}^{(L)}); \quad D_i(u) = \Psi \left( \sum_{s} \phi_s(\mathbf{t}_i^{(s)}) \right)
\end{equation}
where $\mathbf{r}_i \in \mathbb{R}^4$, $\mathbf{t}_i \in \mathbb{R}^3$, and $\mathbf{f}_i \in \mathbb{R}^2$ denote rotation, translation, and focal length, respectively; $u$ denotes the pixel coordinate and $s$ indexes multi-scale features from the backbone.

Training minimizes the joint objective:
\begin{equation}
\mathcal{L} = \mathcal{L}_{action} + \mathcal{L}_{camera} + \mathcal{L}_{depth},
\end{equation}
where $\mathcal{L}_{action}$ regresses the predicted action chunk to the ground-truth trajectory, $\mathcal{L}_{camera}$ is a Huber loss on camera parameters, and $\mathcal{L}_{depth}$ is an uncertainty-weighted depth loss with an additional depth-gradient term.

During inference, the camera and depth heads are bypassed; only the action head decodes the 3D representations. This decoupled design retains the benefits of geometric supervision without the cost of explicit reconstruction, enabling high-frequency control.

\begin{table}[t]
\centering
\caption{\underline{The success rate comparison} on the LIBERO benchmark. The best are highlighted by \textbf{bold}. The results demonstrate VGA’s strong manipulation precision.}
\resizebox{1.0\linewidth}{!}{
\begin{tabular}{lccccc}
\hline
\multicolumn{1}{l|}{Method}                             & Spatial & Object & Goal  & \multicolumn{1}{c|}{Long}                          & Avg \\ \hline
\rowcolor[HTML]{F2F2F2} 
\multicolumn{6}{l}{\cellcolor[HTML]{F2F2F2} \qquad Vision-Language-Action Baselines}                                                                      \\
\multicolumn{1}{l|}{Octo \citeyearpar{team2024octo}}                            & 78.9\%   & 85.7\%  & 84.6\% & \multicolumn{1}{c|}{51.1\%}                         & 75.1\%   \\
\multicolumn{1}{l|}{OpenVLA \citeyearpar{kim2024openvla}}                            & 84.7\%   & 88.4\%  & 79.2\% & \multicolumn{1}{c|}{53.7\%}                         & 76.5\%   \\
\multicolumn{1}{l|}{GR00T-N1 \citeyearpar{bjorck2025gr00t}}                            & 94.4\%   & 97.6\%  & 93.0\% & \multicolumn{1}{c|}{90.6\%}                         & 93.9\%   \\
\multicolumn{1}{l|}{UniVLA \citeyearpar{bu2025univla}}                            & 96.5\%   & 96.8\%  & 95.6\% & \multicolumn{1}{c|}{92.0\%}                         & 95.2\%   \\
\multicolumn{1}{l|}{$\pi_{0}$ \citeyearpar{black2024pi0}}                                & 90.0\%   & 86.0\%  & 95.0\% & \multicolumn{1}{c|}{73.0\%}                         & 86.0\%   \\
\multicolumn{1}{l|}{$\pi_{0.5}$ \citeyearpar{pi05vision}}                                & 98.8\%   & 98.2\%  & 98.0\% & \multicolumn{1}{c|}{92.4\%}                         & 96.9\%   \\
\multicolumn{1}{l|}{GR00T-N1.6 \citeyearpar{bjorck2025gr00t}}                            & 97.7\%   & 98.5\%  & 97.5\% & \multicolumn{1}{c|}{94.4\%}                         & 97.0\%   \\
\multicolumn{1}{l|}{OpenVLA-oft \citeyearpar{kim2025fine}}                            & 97.6\%   & 98.4\%  & 97.9\% & \multicolumn{1}{c|}{94.5\%}                         & 97.1\%   \\
\multicolumn{1}{l|}{VLA-Thinker \citeyearpar{wang2026vla}}                            & 98.7\%   & 99.0\%  & 95.2\% & \multicolumn{1}{c|}{96.9\%}                         & 97.5\%   \\
\rowcolor[HTML]{F2F2F2} 
\multicolumn{6}{l}{\cellcolor[HTML]{F2F2F2} \qquad 3D Vision-Language-Action Baselines}                                                                   \\
\multicolumn{1}{l|}{SpatialVLA \citeyearpar{qu2025spatialvla}}                         & 88.2\%   & 89.9\%  & 78.6\% & \multicolumn{1}{c|}{55.5\%}                         & 78.1\%   \\
\multicolumn{1}{l|}{GeoAwareVLA \citeyearpar{abouzeid2025geoaware}}                         & 95.0\%   & \textbf{100}\%  & 99.0\% & \multicolumn{1}{c|}{93.0\%}                         & 96.8\%   \\
\multicolumn{1}{l|}{GeoVLA \citeyearpar{sun2025geovla}}                         & 98.4\%   & 99.0\%  & 96.6\% & \multicolumn{1}{c|}{96.6\%}                         & 97.7\%   \\
\rowcolor[HTML]{F2F2F2} 
\multicolumn{6}{l}{\cellcolor[HTML]{F2F2F2} \qquad World Action Model Baselines}                                                                          \\
\multicolumn{1}{l|}{UniMimic \citeyearpar{chen2025unifying}}                           & 89.0\%   & 91.0\%  & 85.0\% & \multicolumn{1}{c|}{59.0\%}                         & 81.0\%   \\
\multicolumn{1}{l|}{WorldVLA \citeyearpar{cen2025worldvla}}                           & 87.6\%   & 96.2\%  & 83.4\% & \multicolumn{1}{c|}{60.0\%}                         & 81.8\%   \\
\multicolumn{1}{l|}{UVA \citeyearpar{li2025unified}}                         & -/-\%   & -/-\%  & -/-\% & \multicolumn{1}{c|}{93.0\%}                         & 93.0\%   \\
\multicolumn{1}{l|}{mimic-video \citeyearpar{pai2025mimic}}                           & 94.2\%   & 96.8\%  & 90.6\% & \multicolumn{1}{c|}{-/-\%}                         & 93.9\%   \\
\multicolumn{1}{l|}{Motus \citeyearpar{bi2025motus}}                           & 96.8\%   & 99.8\%  & 96.6\% & \multicolumn{1}{c|}{\textbf{97.6}\%}                         & 97.7\%   \\  \hline
\rowcolor[HTML]{E8F5E9} 
\multicolumn{1}{l|}{\cellcolor[HTML]{E8F5E9}VGA (Ours)} & \textbf{99.0}\%   & 99.6\%  & \textbf{98.6}\% & \multicolumn{1}{c|}{\cellcolor[HTML]{E8F5E9}95.0\%} & \textbf{98.1}\%   \\ \hline
\end{tabular}
}
\label{tab:comparison_libero}
\end{table}

\begin{table}[t]
\centering
\small
\caption{\underline{The success rate comparison} on the Robotwin2.0. Each entry reports Easy/Hard.}
\label{tab:robotwin}
\resizebox{0.47\textwidth}{!}{
\begin{tabular}{@{}cccccc@{}}
\toprule
      & \begin{tabular}[c]{@{}c@{}}adjust\\ bottle\end{tabular} & \begin{tabular}[c]{@{}c@{}}pick\\ dual bottles\end{tabular} & \begin{tabular}[c]{@{}c@{}}turn\\ switch\end{tabular} & \begin{tabular}[c]{@{}c@{}}grab\\ roller\end{tabular} & \begin{tabular}[c]{@{}c@{}}place\\ object basket\end{tabular} \\ \midrule
$\pi_0$   & 99\%/95\%                                                   & 59\%/37\%                                                       & 41\%/42\%                                                 & 98\%/94\%                                                 & 67\%/70\%                                                         \\
X-VLA & \textbf{100}\%/\textbf{99}\%                                                  & 47\%/36\%                                                       & 40\%/\textbf{61}\%                                                 & \textbf{100}\%/\textbf{100}\%                                               & 44\%/39\%                                                         \\ \midrule
\rowcolor[HTML]{E8F5E9} 
VGA   & \textbf{100}\%/\textbf{99}\%                                                  & \textbf{70}\%/\textbf{56}\%                                                       & \textbf{51}\%/56\%                                                 & \textbf{100}\%/\textbf{100}\%                                               & \textbf{81}\%/\textbf{77}\%                                                         \\ \bottomrule
\end{tabular}
}
\end{table}

\begin{table}[t]
\centering
\small
\caption{\underline{The success rate comparison} on the LIBERO-Plus. VGA (col 1) demonstrates robust generalization to unseen viewpoints, lighting shifts, and implicit-goal instructions.}
\label{tab:libero_plus}
\resizebox{0.47\textwidth}{!}{
\begin{tabular}{
c
|>{\columncolor[HTML]{E8F5E9}}c|
ccccc }
\toprule
         & VGA & OpenVLA-oft & $\pi_0$  & WorldVLA & UniVLA & RIPT-VLA  \\ \midrule
light    & \textbf{92.1}\% & 85.8\%        & 79.6\% & 29.4\%     & 59.1\%   & 87.9\% \\
language & \textbf{81.5}\% & \textbf{81.5}\%        & 61.0\% & 44.2\%     & 71.8\%   & 80.1\% \\
camera   & \textbf{86.0}\% & 59.7\%        & 15.8\% & 0.3\%      & 4.3\%    & 58.3\% \\ \bottomrule
\end{tabular}
}
\end{table}

\subsection{Progressive Volumetric Modulation}

Directly attending to the final 3D representations with a single cross-attention layer may underutilize their hierarchical geometric structure.
We therefore introduce Progressive Volumetric Modulation (PVM), a layer-wise module that establishes structured and progressive coupling between multi-scale 3D representations and action decoding.

Specifically, for each layer $l$, PVM ingests three distinct feature sets: the vision-language condition $\mathbf{H}_a^{(l)}$, the action queries condition $\mathbf{H}_b^{(l)}$, and the corresponding decoder latent embedding $\mathbf{h}_{dec}^{(l)}$.
The modulation is executed via a sequential cross-modal transduction sequence.
First, the decoder latent $\mathbf{h}_{dec}^{(l)}$ serves as a query to extract action-relevant context from $\mathbf{H}_b^{(l)}$, followed by a secondary refinement against the spatio-linguistic manifold $\mathbf{H}_a^{(l)}$:
\begin{equation}
\tilde{\mathbf{h}}_{dec}^{(l)} = \text{Attention}(Q=\mathbf{h}_{dec}^{(l)}, K=\mathbf{H}_b^{(l)},V=\mathbf{H}_b^{(l)}),
\end{equation}
\begin{equation}
\mathbf{a}_{dec}^{(l)} = \text{Attention}(Q=\tilde{\mathbf{h}}_{dec}^{(l)}, K=\mathbf{H}_a^{(l)},V=\mathbf{H}_a^{(l)}),
\end{equation}
where the resulting $\mathbf{a}_{dec}^{(l)} \in \mathbb{R}^{C \times D}$ represents the distilled multimodal condition for the current hierarchy.
To ensure a seamless integration with the intrinsic reasoning of the action head, we perform an adaptive manifold alignment.
The modulated feature $\mathbf{a}_{dec}^{(l)}$ is concatenated with the raw updated decoder state $\mathbf{h}_{dec}'^{(l+1)}$, and subsequently projected back to the latent space:
\begin{equation}
\mathbf{h}_{dec}^{(l+1)} = \text{Linear}([\mathbf{h}_{dec}^{\prime(l+1)}, \mathbf{a}_{dec}^{(l)}]),
\end{equation}
By interleaving this dual-stage modulation across all layers, PVM sustains a high-fidelity flow of geometric information into the action generation process.

\subsection{Implementation Details}
The number of transformer layers in the backbone and action head is set as $L=L_a=12$.
We train the model using LoRA~\cite{hu2022lora} to preserve the pretrained capability of the backbone, with a rank of 64. In total, the number of trainable parameters is approximately 500M; a detailed breakdown is provided in the Appendix.
All experiments are conducted on NVIDIA A100-SXM4-80GB GPUs, with the longest training run completed within 60 GPU hours.

\begin{figure}[t]
    \centering
    \includegraphics[width=1.0\linewidth]{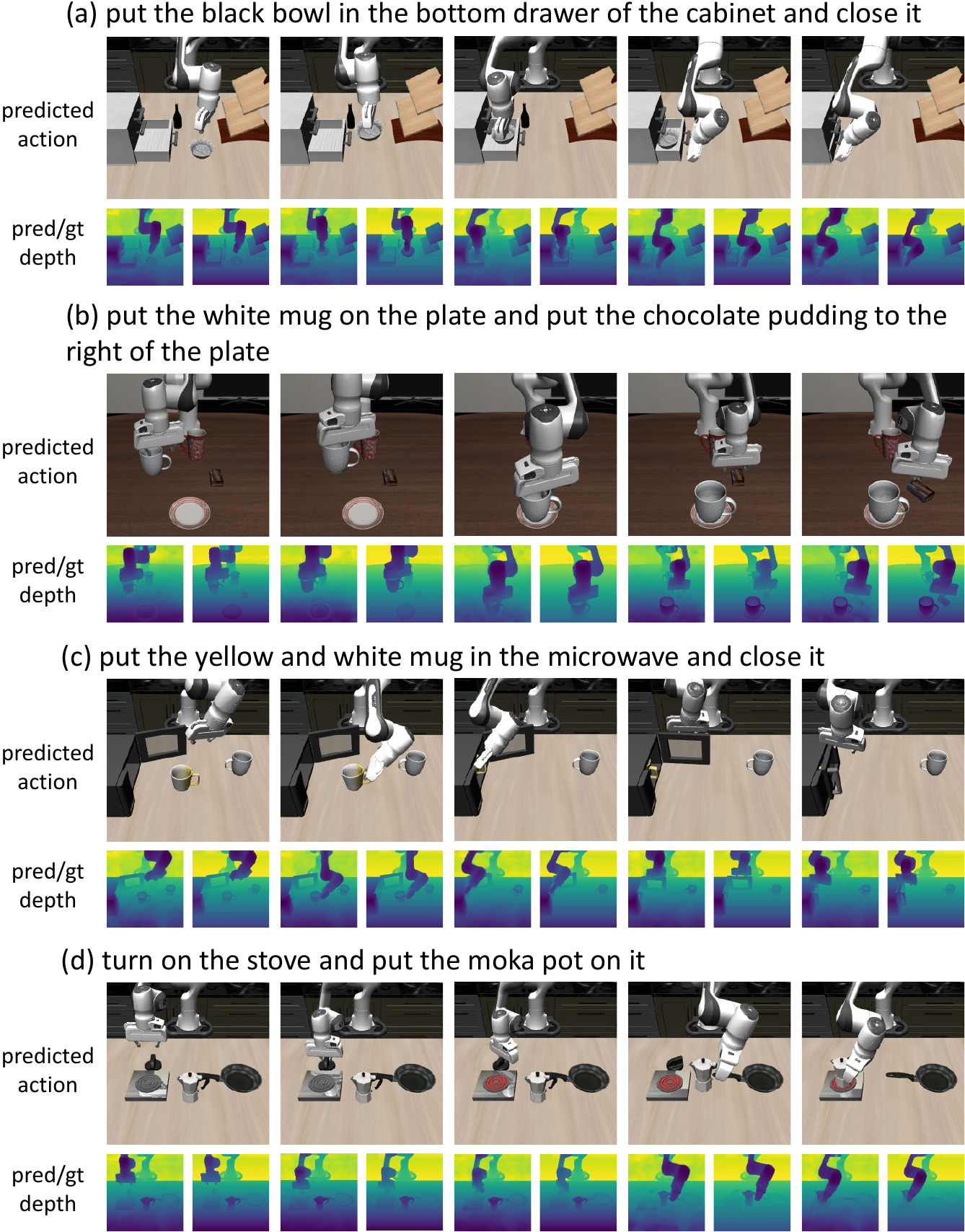}
    \caption{\underline{Simulation rollouts and depth predictions} on the LIBERO benchmark. The results show that VGA achieves precise physical manipulation with precise depth predictions.}
    \label{fig:libero_rollout}
\end{figure}
\section{Simulation Experiment}
This section presents quantitative comparisons between our approach and prominent VLA models in simulated environments, focusing on the following key questions:

\begin{enumerate}
    \renewcommand{\labelenumi}{\textbf{Q\arabic{enumi}.}}
    \item \emph{How does our VGGT-based model perform compared to VLA models built on VLM/Video backbones} (\cref{sec:quantitative_comparison}).
    \item \emph{Are the predicted 3D properties accurate and reliable} (\cref{sec:auxiliary_prediction}).
    \item \emph{What are the individual contributions of each component to the overall performance} (\cref{sec:ablation_study}).
\end{enumerate}

\subsection{Experimental Setup}
\label{sec:simu_setup}

We evaluate VGA on three simulation benchmarks: LIBERO~\cite{liu2023libero}, RoboTwin2.0~\cite{chen2025robotwin}, and LIBERO-Plus~\cite{fei2025libero}, using the average task success rate as the primary metric.

For LIBERO~\cite{liu2023libero}, we follow prior evaluation protocols~\cite{kim2025fine, qu2025spatialvla} and train a separate model for each suite. LIBERO contains four suites, each comprising 10 tasks with approximately 400 training demonstrations. We perform 50 trials per task, totaling 500 rollouts per suite, and obtain ground-truth camera parameters and depth maps directly from the simulator.

RoboTwin2.0~\cite{chen2025robotwin} is a challenging dual-arm benchmark featuring multi-object interactions, diverse scenes, complex semantics, and long-horizon tasks.
We evaluate VGA on five representative tasks and train a separate model for each task. Following prior works~\cite{bi2025motus, li2026causal}, each model is trained using both clean and randomized data.

LIBERO-Plus~\cite{fei2025libero} extends standard LIBERO. It evaluates generalization under controlled variations.
Following its evaluation protocol, we use the LIBERO-trained checkpoints without any further tuning on LIBERO-Plus data. We then evaluate these checkpoints under three perturbation dimensions: lighting, language, and camera.
The language variations assess generalization to unseen instructions requiring commonsense and abstract semantic reasoning, while the camera shifts evaluate zero-shot generalization to substantially different, unseen viewpoints.


\subsection{Quantitative Comparison}
\label{sec:quantitative_comparison}
Quantitative results on LIBERO, RoboTwin 2.0, and LIBERO-Plus are reported in~\cref{tab:comparison_libero,tab:robotwin,tab:libero_plus}, respectively, while qualitative LIBERO rollouts are presented in~\cref{fig:libero_rollout}.

\textbf{LIBERO.} On the standard LIBERO benchmark, VGA achieves the highest average success rate of 98.1\%, outperforming the strong VLA baselines $\pi_{0.5}$ and OpenVLA-OFT by 1.2 and 1.0 percentage points, respectively. It also surpasses the 3D-VLA baselines SpatialVLA, GeoAwareVLA, and GeoVLA by 20.0, 1.3, and 0.4 percentage points, respectively, and slightly exceeds the strongest WAM baseline, Motus, by 0.4 percentage points. These results suggest that a pretrained 3D world model can serve as an effective and competitive backbone for robotic manipulation.
Across individual suites, VGA performs particularly well on Spatial, Object, and Goal, highlighting its strength in tasks requiring spatial reasoning and precise manipulation. Its 95.0\% success rate on LIBERO-Long remains slightly below the strongest baselines, indicating that its advantage is more evident in spatially demanding tasks than in long-horizon settings.

\begin{table*}[t]
\caption{\underline{Ablation study} across all LIBERO suites. We evaluate the effects of PVM, joint 3D supervision, VGGT pretraining, and training strategy. The results demonstrate the contribution of each design component.}
\begin{tabular}{l|cccc|cccc|c}
\hline
Method                       & PVM & \begin{tabular}[c]{@{}c@{}}Training\\ Objective\end{tabular} & Initialization  & \begin{tabular}[c]{@{}c@{}}Training\\ Strategy\end{tabular} & Spatial & Object & Goal & Long & \cellcolor[HTML]{FFFFFF}Avg  \\ \hline
VGA \small{\emph{w/o PVM}}                  & -   & Action + 3D                                                  & VGGT-Pretrained & LoRA                                                        & 96.2    & 99.0   & 97.4 & 90.0 & \cellcolor[HTML]{FFFFFF}95.7 \\
VGA \small{\emph{w/o joint-training}}       & +   & Action                                                       & VGGT-Pretrained & LoRA                                                        & 97.8    & 99.2   & 97.8 & 94.0 & \cellcolor[HTML]{FFFFFF}97.2 \\
VGA \small{\emph{zeroinit, full-parameter}} & +   & Action + 3D                                                  & Random          & Full                                                        & 86.6    & 90.4   & 88.4 & 81.0 & \cellcolor[HTML]{FFFFFF}86.6 \\
VGA \small{\emph{zeroinit, lora}}           & +   & Action + 3D                                                  & Random          & LoRA                                                        & 8.4     & 10.8   & 6.8  & 0.0  & \cellcolor[HTML]{FFFFFF}6.4  \\
VGA \small{\emph{full-parameter}}           & +   & Action + 3D                                                  & VGGT-Pretrained & Full                                                        & 92.6    & 92.0   & 88.2 & 75.4 & \cellcolor[HTML]{FFFFFF}87.1 \\
\rowcolor[HTML]{F2F2F2} 
VGA                          & +   & Action + 3D                                                  & VGGT-Pretrained & LoRA                                                        & \textbf{99.0}    & \textbf{99.6}   & \textbf{98.6} & \textbf{95.0} & \textbf{98.1}                         \\ \hline
\end{tabular}
\label{tab:ablation_study_add}
\end{table*}

\begin{figure}[t]
    \centering
    \includegraphics[width=0.95\linewidth]{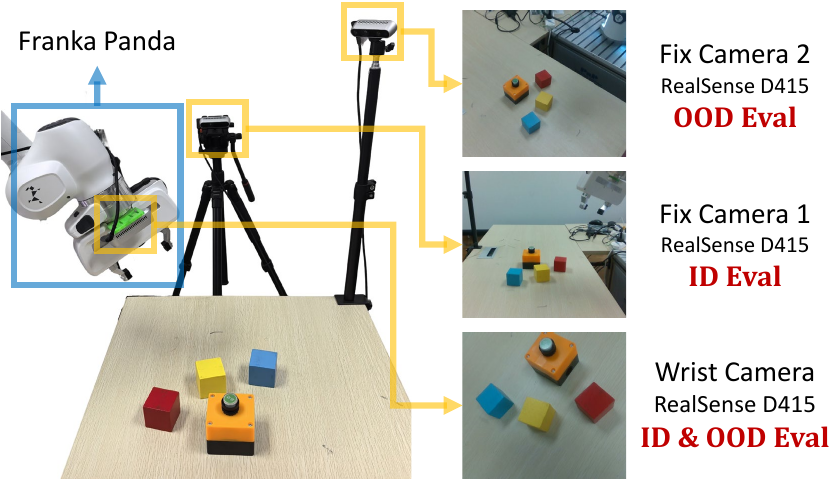}
    \caption{\underline{Configuration of real world evaluation.} The platform is equipped with one wrist camera and two fixed cameras. The fixed cameras are used for in-distribution and out-of-distribution evaluation, respectively.}
    \label{fig:realworld_config}
\end{figure}

\textbf{RoboTwin2.0.} As shown in~\cref{tab:robotwin}, VGA achieves average success rates of 80.4\% and 77.6\% under the Easy and Hard settings, respectively, compared with 72.8\%/67.6\% for $\pi_0$ and 66.2\%/67.0\% for X-VLA. Given the challenging nature of RoboTwin 2.0, which involves multiple objects, diverse scenes, and complex bimanual interactions, these results suggest that VGA maintains precise spatial control even in challenging manipulation scenarios.

\textbf{LIBERO-Plus.} Without any additional training on LIBERO-Plus data, VGA achieves success rates of 92.1\%, 81.5\%, and 86.0\% under lighting, language, and camera variations, respectively. It achieves the best performance under lighting and camera variations while matching OpenVLA-OFT under language variations. Most notably, under large camera-viewpoint shifts, VGA surpasses the second-best OpenVLA-OFT by 26.3 percentage points. This result provides further evidence of VGA’s zero-shot generalization to substantially different, unseen viewpoints.

Overall, the results highlight VGA’s strength in precise spatial control and support pretrained 3D world models as a promising alternative backbone for robotic manipulation.

\subsection{Quality of Auxiliary 3D Prediction}
\label{sec:auxiliary_prediction}

\cref{fig:libero_rollout} compares the predicted and ground-truth depth maps alongside action rollouts. VGA produces accurate depth estimates, particularly for manipulated objects, indicating that the learned representations preserve task-relevant geometry and facilitate alignment between scene representations and 3D physical actions.

\subsection{Ablation Study}
\label{sec:ablation_study}

We conduct ablations across all four LIBERO suites to evaluate PVM, joint 3D supervision, VGGT pretraining, and training strategy. As shown in~\cref{tab:ablation_study_add}, removing PVM lowers the average success rate by 2.4 percentage points and by 5.0 points on LIBERO-Long, supporting multi-scale interaction between 3D representations and action queries. Removing auxiliary 3D supervision reduces the average from 98.1\% to 97.2\%, suggesting that joint training further improves action-relevant representations.

The results also show a strong interaction between initialization and training strategy. With VGGT pretraining, LoRA outperforms full-parameter training (98.1\% vs. 87.1\%); from random initialization, however, LoRA achieves only 6.4\%, compared with 86.6\% for full-parameter training. Thus, LoRA’s advantage stems not from parameter-efficient tuning alone, but from preserving and leveraging the pretrained VGGT’s structured 3D priors.

\section{Real World Experiment}

\begin{table*}[t]
\caption{\underline{Real world evaluation} of fine-grained manipulation. We report success rates under ID and OOD camera viewpoints. VGA achieves the highest average success rate in the OOD setting, demonstrating strong zero-shot cross-view generalization.}
\begin{tabular}{@{}lccc
>{\columncolor[HTML]{F2F2F2}}c ccc
>{\columncolor[HTML]{F2F2F2}}c @{}}
\toprule
\multirow{2}{*}{Method} & \multicolumn{4}{c}{In-Distribution Evaluation}                          & \multicolumn{4}{c}{Out-of-Distribution Evaluation}                         \\ \cmidrule(l){2-9} 
                        & \emph{Pick Cube} & \emph{Press Button} & \emph{Stack Cube} & Average & \emph{Pick Cube} & \emph{Press Button} & \emph{Stack Cube} & Average \\ \midrule
ACT                     & 40\%     & 40\%        & 30\%      & 37\%   & 10\%     & 10\%        & 0\%      & 7\%   \\
OpenVLA                 & 30\%     & 25\%        & 0\%      & 18\%   & 5\%     & 5\%        & 0\%      & 3\%   \\
$\pi_{0.5}$                    & \textbf{85}\%     & \textbf{85}\%        & \textbf{60}\%      & \textbf{77}\%   & 50\%     & 55\%        & \textbf{50}\%      & 52\%   \\
VGA (Ours)              & 80\%     & \textbf{85}\%        & \textbf{60}\%      & 75\%   & \textbf{70}\%     & \textbf{65}\%        & 40\%      & \textbf{58}\%   \\ \bottomrule
\end{tabular}
\label{tab:real_world_performance}
\end{table*}



\begin{figure*}[t]
    \centering
    \includegraphics[width=1.0\linewidth]{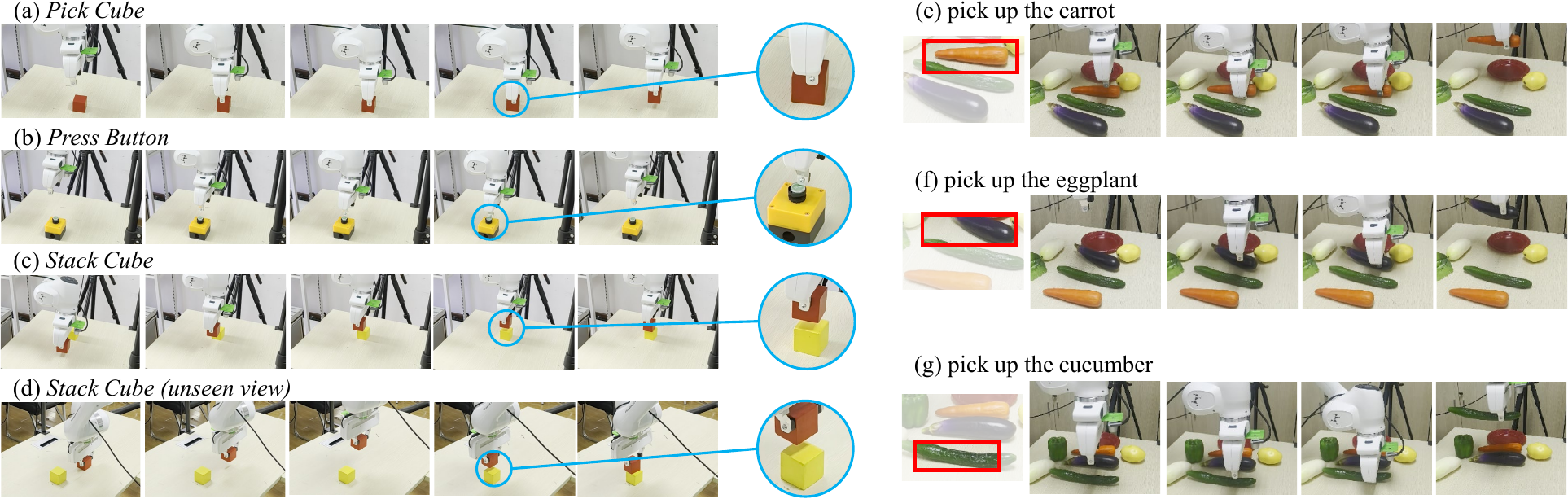}
    \caption{\underline{Fine-grained and language-grounded real-world manipulation.} (a–d) show VGA performing fine-grained manipulation under in-distribution and unseen camera views. (e–g) show VGA following language instructions to grasp the specified target among three visually similar objects under varying layouts. These results demonstrate VGA’s fine-grained control across ID and OOD views and accurate language following, highlighting its generalization and real-world deployment potential.}
    \label{fig:realworld_rollout}
\end{figure*}

We conduct a series of real-world robot manipulation experiments to answer three key questions:

\begin{enumerate}
    \renewcommand{\labelenumi}{\textbf{Q\arabic{enumi}.}}
    \setcounter{enumi}{3}
    \item \emph{Can VGA achieve reliable task execution in real-world settings} (\cref{sec:In_Distribution_Evaluation}).
    \item \emph{Can VGA generalize zero-shot to unseen camera viewpoints} (\cref{sec:Out_of_Distribution_Generalization}).
    \item \emph{Can VGA perform language-grounded real-world manipulation} (\cref{sec:Semantic_Complex_Manipulation}).
\end{enumerate}

\subsection{Experimental Setup}
\label{sec:Experimental_Setup}

Our setup consists of a Franka Panda robot and three RealSense D415 cameras: one wrist camera and two fixed cameras providing different external viewpoints (\cref{fig:realworld_config}). One fixed camera is used during data collection and in-distribution evaluation, whereas the other provides an unseen viewpoint for out-of-distribution evaluation.

\begin{table}[t]
\centering
\caption{\underline{Real-world evaluation} of language-grounded manipulation. We report success rates for grasping the instructed vegetable in unseen layouts containing all three visually similar objects.}
\label{tab:language_grasping}
\begin{tabular}{@{}l|ccc|c@{}}
\toprule
Method & pick carrot & pick eggplant & pick cucumber & avg \\ \midrule
$\pi_{0.5}$     & 75\%          & 75\%            & 80\%            & 76.7\%  \\
\rowcolor[HTML]{E8F5E9} 
VGA    & 80\%          & 80\%            & 80\%            & 80.0\%  \\ \bottomrule
\end{tabular}
\end{table}

Following prior real-world evaluation protocols~\cite{xu2025a0, zhang2025dreamvla, wang2025vla}, we consider fine-grained and language-grounded manipulation. The fine-grained tasks include \emph{Pick Cube}, \emph{Press Button}, and \emph{Stack Cube}. We collect 80–100 teleoperated demonstrations per task and conduct 20 trials per task under both the training viewpoint (ID) and an unseen viewpoint (OOD). For language-grounded manipulation, we collect 60 demonstrations for each of three visually similar targets—carrot, eggplant, and cucumber—with only the target object present during training. During evaluation, all three objects are placed together in unseen layouts, and each target instruction is tested over 20 trials.

These experiments use action supervision only because acquiring dense real-world 3D annotations is costly.
We compare VGA with ACT~\cite{zhao2023learning}, OpenVLA~\cite{kim2024openvla}, and $\pi_{0.5}$~\cite{pi05vision} on the fine-grained manipulation tasks, and with $\pi_{0.5}$~\cite{pi05vision} on language-grounded manipulation. Detailed baseline configurations are provided in Appendix.

\subsection{In-Distribution Evaluation}
\label{sec:In_Distribution_Evaluation}
We first evaluate the three fine-grained tasks under the training viewpoint. As shown in~\cref{tab:real_world_performance} and~\cref{fig:realworld_rollout}, VGA achieves an average success rate of 75\%, outperforming ACT (37\%) and OpenVLA (18\%). It remains comparable to $\pi_{0.5}$ (77\%), matching its performance on Press Button (85\%) and Stack Cube (60\%), while trailing by 5 percentage points on Pick Cube (80\% vs. 85\%). These results suggest that a pretrained vision-geometry backbone can effectively support fine-grained real-world manipulation.

\subsection{Out-of-Distribution Generalization}
\label{sec:Out_of_Distribution_Generalization}

To evaluate zero-shot cross-view generalization, we use the same demonstrations collected with the wrist camera and camera-1, but replace camera-1 with the unseen camera-2 during evaluation (\cref{fig:realworld_config}), without further training. This setting tests whether the learned policies can transfer across camera viewpoints.

As shown in~\cref{tab:real_world_performance}, ACT and OpenVLA achieve average success rates of only 7\% and 3\%, respectively, while $\pi_{0.5}$ reaches 52\%. VGA achieves 58\%, surpassing $\pi_{0.5}$ by 6 percentage points.
These results highlight a key advantage of our approach. By leveraging a pretrained 3D world model, VGA learns structured representations that capture the underlying scene geometry and action-relevant spatial relationships. During imitation learning, the model learns a direct mapping from these geometry-oriented representations to future actions, rather than relying solely on viewpoint-dependent 2D visual patterns. This enables VGA to better preserve spatial relationships when the viewpoint changes and generate appropriate actions under unseen views. The results therefore suggest that pretrained 3D representations provide a useful basis for improving zero-shot cross-view generalization in real-world manipulation.

\subsection{Language-Grounded Manipulation}
\label{sec:Semantic_Complex_Manipulation}


We evaluate whether VGA can perform language grounding in real-world manipulation. The scene contains three visually similar vegetables with different semantic identities, and the language instruction specifies which object to grasp. Successfully completing the task requires distinguishing between the object categories, associating the instruction with the correct object, and locating the specified target in a multi-object scene before executing the grasp. These requirements are related to prior work on object-centric manipulation and object-oriented embodied reasoning~\cite{shridhar2022cliport,jiang2023vima,chen2026oowm}.

As shown in~\cref{tab:language_grasping}, VGA achieves success rates comparable to $\pi_{0.5}$ across all three language-grounded tasks, reaching an average success rate of 80\%. The real-world rollouts in~\cref{fig:realworld_rollout} (e)-(g) further show that VGA consistently identifies and grasps the instructed object across different target positions and spatial layouts.

These results show that VGA can accurately follow language instructions, distinguish visually similar objects with different semantics, and locate the requested target in multi-object scenes. This capability further supports the use of a pretrained 3D world model as a viable backbone for robotic manipulation.
\section{Conclusion}
In this work, we formalize robotic manipulation fundamentally as a vision-to-geometry mapping ($f(v) \rightarrow G$) and present the Vision-Geometry-Action (VGA) model. By shifting the paradigm away from conventional language and video models toward a native 3D vision-geometry backbone (VGGT), VGA seamlessly bridges the gap between visual observations and spatially grounded physical actions. Extensive experiments strongly validate this geometry-first design.
Across LIBERO, RoboTwin2.0, and LIBERO-Plus, VGA outperforms representative VLA, 3D-VLA, and WAM baselines, demonstrating precise manipulation and robust generalization in simulation. Moreover, without relying on additional 3D sensors, VGA achieves superior results compared with 3D-aware VLA approaches such as GeoVLA, bypassing the 3D-2D-3D information bottleneck. In real-world deployments, VGA exhibits strong robustness, surpassing $\pi_{0.5}$ by 6 percentage points under unseen camera viewpoints and accurately following language instructions for target grasping. Ablation studies confirm the contributions of Progressive Volumetric Modulation and joint training, while accurate 3D property predictions show that VGA preserves geometrically consistent scene representations. Overall, these findings demonstrate that vision-geometry backbones, rather than language or video priors, offer a promising direction for generalizable robotic manipulation.

\begin{acks}
This work was supported in part by National High-Level Young Talent Program (Grant 2025HY00260104), in part by the Fundamental Research Funds for Higher Education Institutions allocated to Sun Yat-sen University (Grant 25hytd007), in part by the Guangdong Provincial High-Level Young Talent Program (Grant 2025HYSPT0707), in part by the Tuoyuan Fund (Grant HT-99982025-0564), in part by the Faculty Start-up Research Fund (Grant 67000-12255002), in part by Scientific Research and Innovation Development Project (Grant 67000-12261023), and in part by the Huawei Strategic Research Institute Talent Fund.
\end{acks}

\clearpage

\bibliographystyle{ACM-Reference-Format}
\bibliography{sample-base}

\clearpage
\appendix

\section{Additional Experiments}

\subsection{Additional Real World Experiment}
In this section, we present an additional real-world experiment to further demonstrate the reliability of VGA in practical deployment, especially under different language conditions.
While the main paper evaluates standard manipulation tasks such as Pick Cube, Press Button, Stack Cube, and cross-view generalization, here we focus on a more challenging language-conditioned grasping scenario with visually similar objects.

\begin{figure}[t]
    \centering
    \includegraphics[width=1.0\linewidth]{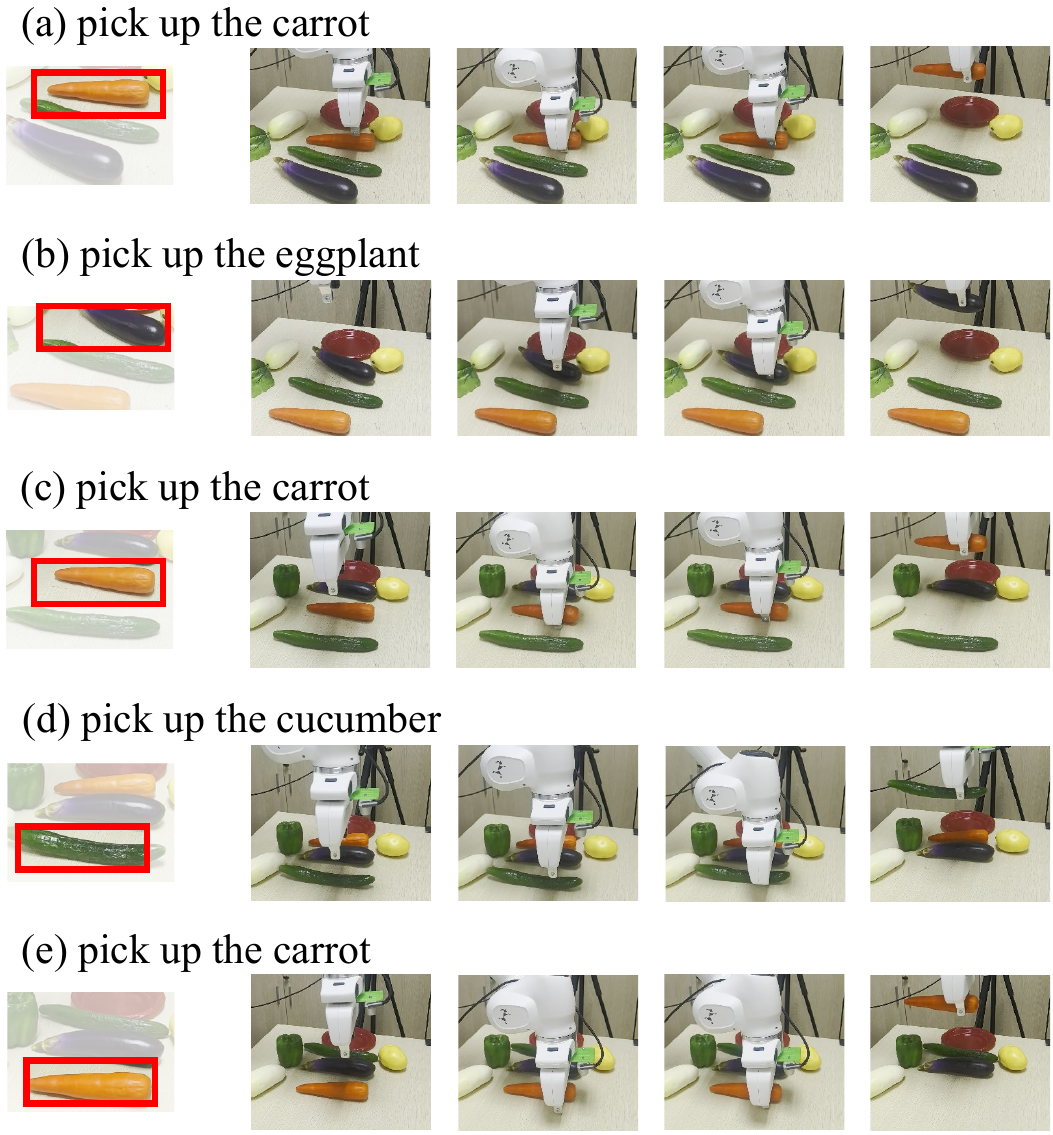}
    \caption{\underline{Language-conditioned grasping} under varying layouts. This figure presents the results of real-world grasping with three visually similar objects arranged in different layouts. Each row corresponds to a different spatial configuration, and the robot is instructed to pick a target object. VGA consistently identifies and grasps the correct object regardless of its position, demonstrating robust language grounding and reliable real-world manipulation performance.}
    \label{fig:add_real_world_rollout}
\end{figure}

Specifically, we select three vegetables with similar shapes as target objects, namely cucumber, carrot, and eggplant. These objects are placed on the table in arbitrary orders under different layout configurations, as shown in~\cref{fig:add_real_world_rollout}. The robot is then instructed via language to pick up a specific object. We train VGA using 60 demonstration trajectories and evaluate its ability to follow language instructions under diverse spatial arrangements.

The results show that VGA consistently follows the language command and successfully grasps the correct object across different layouts. For example, in the first row of~\cref{fig:add_real_world_rollout}, the carrot is positioned on the left side relative to the robot, and VGA successfully picks it up when instructed. In the third and fifth rows, where the carrot appears in the middle and on the right side, respectively, VGA continues to correctly execute the command. Similar behavior is observed for the cucumber and eggplant. These results indicate that VGA can robustly ground language instructions to the correct visual targets, even when objects are visually similar and spatially rearranged. This capability is enabled by the Qwen-GTE language encoder, which provides strong semantic representations for aligning language with visual observations.

\subsection{Ablation on LoRA Rank}
\begin{figure}[t]
    \centering
    \includegraphics[width=1.0\linewidth]{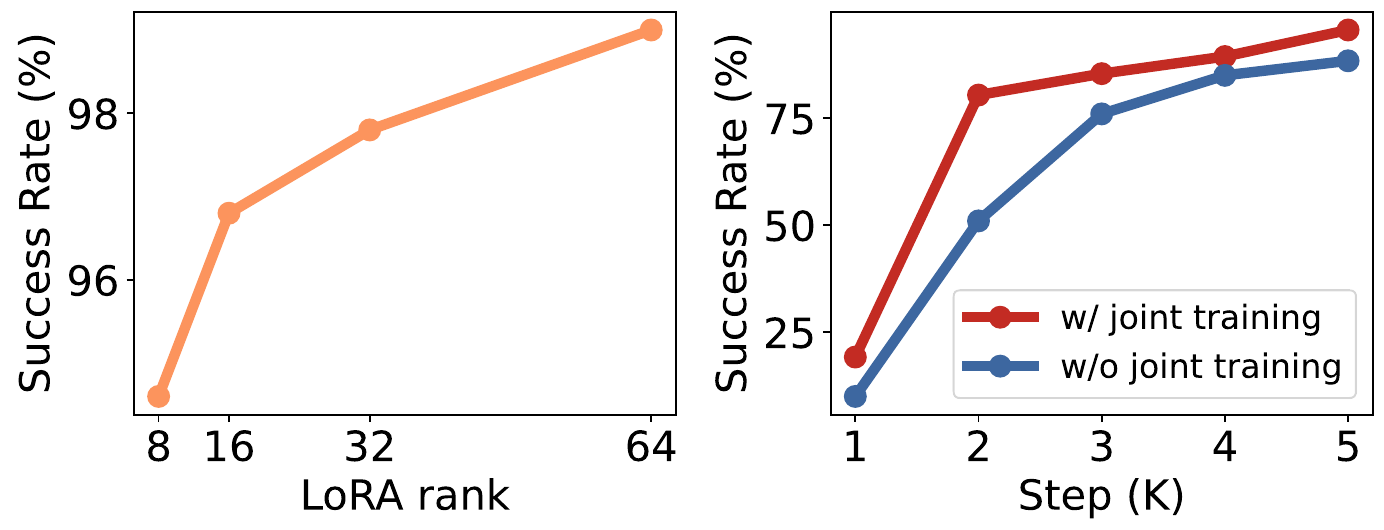}
    \caption{\underline{Impact of joint training and LoRA rank.} Part (a) presents the ablation study on LoRA rank. The results plateau around rank 64. Part (b) compares the convergence speed between models trained with and without joint training, evaluated using checkpoints from 1K to 5K steps on LIBERO-Spatial. Joint training leads to faster convergence and improved early-stage performance, indicating better data efficiency.}
    \label{fig:data_effi_and_ablation}
\end{figure}

We study the impact of the LoRA rank on model performance. As shown in~\cref{fig:data_effi_and_ablation}-(a), increasing the rank generally improves performance, while the gains gradually diminish as the rank becomes larger. Based on this observation, we adopt a LoRA rank of 64 in our final model to ensure stable and strong performance.


\subsection{Data Efficiency with Joint Training}

In this section, we evaluate the effect of joint training on data efficiency. Here, joint training refers to jointly optimizing both 3D property supervision and action supervision, while the variant without joint training is trained using action supervision only. In the main paper, we have already shown the impact of joint training on final performance in~\cref{tab:ablation_study_add}. Building upon this, we further examine the convergence behavior of the two variants to better understand their training dynamics.
Specifically, we compare models trained with and without joint training by measuring the success rate of intermediate checkpoints from 1K to 5K training steps on the LIBERO-Spatial benchmark, as shown in~\cref{fig:data_effi_and_ablation}-(b). The results show that joint training consistently achieves higher success rates at earlier stages of training, indicating a faster convergence process. This suggests that incorporating 3D property supervision facilitates more effective cross-modal interaction, enabling the model to capture the underlying action patterns more efficiently and thereby improving data efficiency.

\subsection{Inference Latency}

In this section, we report the inference latency of our method and compare it with several representative VLA approaches. The results are summarized in~\cref{tab:infer_speed}. For the baseline methods, the reported latency is obtained from their original papers or reliable reproductions~\cite{li2025unified} to ensure fair comparison.
Our method achieves a low inference latency of approximately 0.1 seconds, without applying any hardware-specific optimization techniques. This performance significantly outperforms prior VLA methods such as OpenVLA~\cite{kim2024openvla} and TraceVLA~\cite{zheng2024tracevla}, demonstrating the efficiency of our design. In practice, this corresponds to an inference frequency of around 10 Hz, indicating that our model is well-suited for real-time robotic control scenarios.

\begin{table}[t]
    \centering
    \setlength{\tabcolsep}{12pt}
    \caption{\underline{Inference latency.} VGA achieves low latency of approximately 100 ms (\textasciitilde10 Hz), demonstrating its strong efficiency for real-time robotic control.}
    \begin{tabular}{ccc}
      \toprule
      Method & Latency$\downarrow$ & Frequency$\uparrow$
      \\
      \midrule
RT-2~\cite{zitkovich2023rt} & \textasciitilde200 ms & \textasciitilde5 Hz \\
OpenVLA~\cite{kim2024openvla} & \textasciitilde160 ms & \textasciitilde6 Hz \\
$\pi_{0}$~\cite{black2024pi0} & \textasciitilde100 ms & \textasciitilde10 Hz \\
UVA~\cite{li2025unified} & \textasciitilde230 ms & \textasciitilde4 Hz \\
WorldVLA~\cite{cen2025worldvla} & \textasciitilde330 ms & \textasciitilde3 Hz  \\
TraceVLA~\cite{zheng2024tracevla} & \textasciitilde160 ms & \textasciitilde6 Hz  \\
GraspVLA~\cite{deng2025graspvla} & \textasciitilde200 ms & \textasciitilde5 Hz  \\
      \midrule
VGA (Ours) & \textasciitilde100 ms & \textasciitilde10 Hz \\
      \bottomrule
    \end{tabular}
    \label{tab:infer_speed}
\end{table}

\begin{figure}[b]
    \centering
    \includegraphics[width=1.0\linewidth]{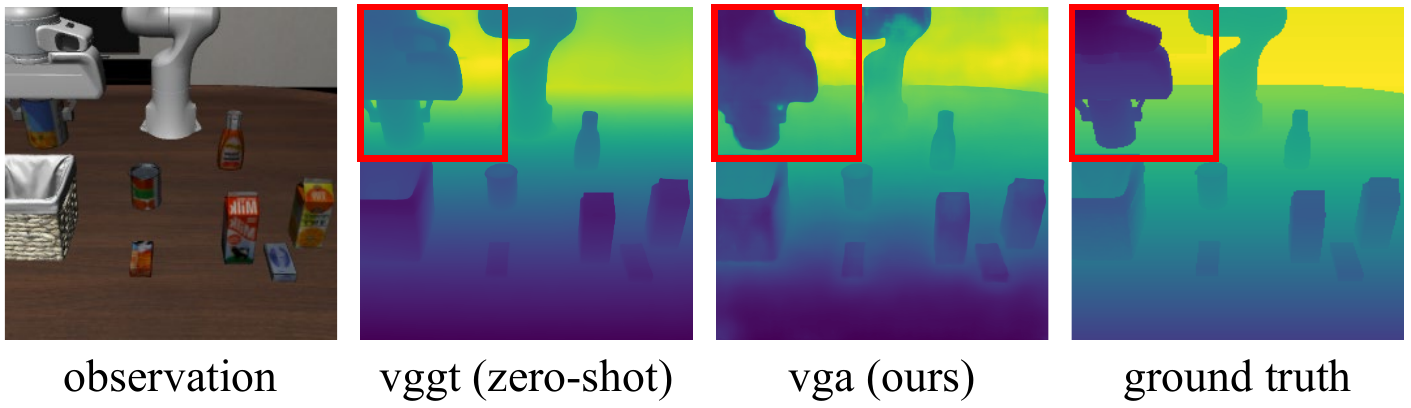}
    \caption{\underline{Depth prediction comparison} between VGGT and VGA. Red boxes highlight regions where VGA significantly differs from VGGT. VGA produces more accurate depth estimates, particularly for the robot gripper, demonstrating the benefit of joint training in improving spatial understanding and depth prediction accuracy.}
    \label{fig:qualitative_depth_compare}
\end{figure}

\subsection{Qualitative Analysis}
In this section, we provide a qualitative comparison of depth prediction results to better understand the effect of joint training. As shown in~\cref{fig:qualitative_depth_compare}, we compare the predicted depth maps from our VGA model and the zero-shot VGGT model~\cite{wang2025vggt} under the same input observations. While VGGT provides reasonable global structure, it fails to accurately infer the depth of certain regions due to the lack of task-specific supervision. In particular, as highlighted by the red boxes, VGGT incorrectly assigns similar depth values to the robot gripper and the robot body, likely due to their similar visual appearance and the absence of additional contextual cues.
In contrast, VGA produces more accurate depth predictions in these challenging regions. Through joint training with both 3D property supervision and action supervision, the model learns to better capture the spatial relationships between the robot and the environment. For example, VGA correctly predicts that the gripper is closer in depth to the target basket, which is consistent with the underlying manipulation objective. This qualitative result demonstrates that cross-modal interaction during joint training improves the model’s ability to infer precise 3D structure, leading to more accurate and task-consistent depth estimation.

\section{Implementation Details}

\subsection{Model Architecture Details}
In this section, we describe the architectural details of VGA. To maintain consistency with the original VGGT~\cite{wang2025vggt} design, our backbone consists of 12 transformer layers, evenly divided into Global Attention and Local Attention blocks. Building on this structure, the action head is designed with the same 12-layer architecture, allowing intermediate representations from the backbone to be effectively propagated into the action prediction module in a manner analogous to KV-cache reuse, thereby improving information flow across stages.

The action query is defined with a length of 8, which is aligned with the chunk size used during training and inference. For visual inputs, the number of camera tokens is set to 16, directly following the default configuration of VGGT. The visual observations are first encoded by a DINO-based encoder, and the resulting features are further projected into the latent space through an MLP before being fed into the transformer backbone. For language input, we employ Qwen-GTE-1.5B~\cite{li2023towards}, a general-purpose text embedding model from the Qwen family that is designed to produce high-quality semantic representations for diverse language understanding tasks. The encoded token-level representations are similarly projected into the latent space via an MLP, and the final valid token is selected as the language token to interact with the rest of the model.

During training, we adopt a parameter-efficient tuning strategy using LoRA~\cite{hu2022lora}. Specifically, LoRA is applied only to the linear layers within the transformer blocks, including the projection layers for query, key, and value, as well as the final output projection layers. This design limits the number of trainable parameters while preserving the overall model capacity.

\subsection{VGA Parameter Composition}
This section provides a detailed breakdown of the parameter composition of VGA, with the full statistics reported in~\cref{tab:vga_param_count}. Overall, VGA maintains a relatively lightweight design compared to existing approaches. Moreover, during training, we adopt a LoRA-based adaptation strategy, which significantly reduces the number of trainable parameters by restricting updates to a small set of low-rank components. This design enables efficient optimization while preserving the expressiveness of the underlying model.

\begin{table}[t]
    \centering
    \setlength{\tabcolsep}{12pt}
    \caption{\underline{VGA parameter count.} This table shows the detailed breakdown of the total and trainable parameters across different modules in VGA. By leveraging the pretrained 3D prior of VGGT, VGA significantly reduces the number of trainable parameters, enabling an efficient and expressive design.}
    \begin{tabular}{ccc}
      \toprule
      Module & Parameters & Learnable \\
      \midrule
Vision Encoder & 730.9M & 0.0M (frozen) \\
Language Encoder & 1543.3M & 0.0M (frozen) \\
Proprio Encoder & 1.1M & 1.1M \\
Vision Projector & 27.6M & 27.6M \\
Language Projector & 2.2M & 2.2M \\
Transformer Backbone & 987.0M & 214.1M \\
Action Head & 284.5M & 284.5M \\
Depth Head & 32.7M & 32.7M \\
      \midrule
Total & 3609.3M & 562.2M \\
      \bottomrule
    \end{tabular}
    \label{tab:vga_param_count}
\end{table}

\subsection{Hyperparameters and Training Details}

In this section, we provide the training details for VGA. The full set of hyperparameters is reported in~\cref{tab:vga_hyperparameters}. For all simulation experiments, the model consistently converges within 120K training steps, and we select the checkpoint that achieves the best performance during training. For real-world experiments, convergence is reached within 20K steps, and similarly, the best-performing checkpoint is used for evaluation.

In simulation, training is performed with joint supervision on both 3D properties and actions, where the ground-truth 3D property labels are directly obtained from the simulator backend. In contrast, for real-world experiments, although the RealSense D415 camera can provide depth observations, we find that these measurements are often noisy and difficult to accurately calibrate. As a result, we rely solely on action supervision in the real-world setting, without explicit 3D property supervision. Instead, the model leverages the pretrained 3D prior from VGGT to provide meaningful 3D representations and support cross-view generalization. As shown in~\cref{tab:real_world_performance} of the main paper, this design remains effective in practice, demonstrating that the VGGT prior alone is sufficient to enable strong generalization performance even without additional 3D supervision in real-world scenarios.

\subsection{Training Details for Baseline}

In this section, we provide additional training details for the baseline methods used in the real-world experiments. The complete set of hyperparameters for each method is reported in \cref{tab:act_hyperparameters_aloha} (ACT), \cref{tab:openvla_oft_hyperparameters_aloha} (OpenVLA), and \cref{tab:pi0_hyperparameters_aloha} ($\pi_{0.5}$).
All baselines are trained following their standard configurations, and model selection is based on convergence behavior observed during training.
For ACT, we select the checkpoint at 5K training steps, where the model has already converged and demonstrates stable performance. For OpenVLA, although the training loss largely converges around 30K steps, we observe more stable performance at 50K steps in our experiments, and thus adopt the 50K checkpoint. For $\pi_{0.5}$, we use the checkpoint at 20K steps, where the model has reached convergence and maintains stable performance.
This setup ensures that each method is evaluated after reaching a sufficiently stable training stage while respecting their differing optimization dynamics.

\section{VGGT Architecture}

In this work, we leverage the Visual Geometry Grounded Transformer (VGGT)~\cite{wang2025vggt} as the spatial perception foundation for robotic manipulation.
VGGT is a unified feed-forward framework designed to infer comprehensive 3D scene attributes from a flexible number of input images.
Specifically, given a sequence of $N$ RGB observations $\mathcal{I} = \{I_{1}, I_{2}, \dots, I_{N}\}$, the model maps these inputs into a high-dimensional representation space to jointly estimate camera parameters $g_i$, depth maps $D_i$, viewpoint-invariant point maps $P_i$, and dense tracking features $T_i$ for each frame.

The core architecture of VGGT is a large-scale Transformer backbone that processes image tokens with minimal 3D-inductive biases. Each image $I_i$ is first partitioned into $K$ patches and projected into a sequence of initial tokens $\{t_{i,k}^{(0)}\}_{k=1}^K$, where $t_{i,k} \in \mathbb{R}^C$. To facilitate geometric reasoning across views, the backbone employs an Alternating-Attention (AA) mechanism across $L$ transformer blocks. Each block consists of two successive attention stages: a frame-wise local attention stage and a cross frame global attention stage. Let $t_{i,k}^{(2l)}$ denote the token at index $k$ of image $i$ entering the $l$-th AA block. The evolution of the tokens is defined by:
\begin{equation}
t_{i,k}^{(2l+1)}=\sum_{m=1}^K{\mathrm{softmax} \left( \frac{(W_{Q}t_{i,k}^{(2l)})^{\top}(W_{K}t_{i,m}^{(2l)})}{\sqrt{d}} \right) W_{V}t_{i,m}^{(2l)}}
\end{equation}
\begin{equation}
t_{i,k}^{(2l)}=\sum_{j=1}^N{\sum_{m=1}^K{\mathrm{softmax} \left( \frac{(W_{Q}t_{i,k}^{(2l+1)})^{\top}(W_{K}t_{j,m}^{(2l+1)})}{\sqrt{d}} \right) W_{V}t_{j,m}^{(2l+1)}}}
\end{equation}
where $W_Q, W_K, W_V$ are the query, key, and value projection matrices. In the frame-wise stage (layer $2l$), attention is restricted to tokens within the same image $i$, capturing local intra-image structure. In the global stage (layer $2l+1$), every token attends to all tokens across all $N$ frames, enabling the model to integrate cross-view geometric constraints and establish a global spatial context.

\begin{table*}
\centering
\caption{VGA hyperparameters for both LIBERO and real-world experiments.}
\label{tab:vga_hyperparameters}
\begin{tabular}{ll}
\toprule
hyperparameter & value \\
\midrule
\# GPUs & 4 x NVIDIA A100-SXM4-80GB GPU \\
learning rate (LR) & 2e-4 \\
total batch size & 32 \\
training strategy & LoRA tuning, LoRA rank = 64 \\
input images & 1 third-person camera image, 1 wrist-mounted camera image \\
input image size & 224 x 224 px \\
use observation history & no (use single-step inputs) \\
action chunk size & 8 steps (predict 8, execute all 8 open-loop at test time) \\
use proprio (robot state) & yes \\
use FiLM & yes \\
\# trainable parameters & 500M total \\
image augmentations & 90\% random crops, color jitter: \\
& \;\; \texttt{random\_resized\_crop=dict(scale=[0.9, 0.9], ratio=[1.0, 1.0])} \\
& \;\; \texttt{random\_brightness=[0.2]} \\
& \;\; \texttt{random\_contrast=[0.8, 1.2]} \\
& \;\; \texttt{random\_saturation=[0.8, 1.2]} \\
& \;\; \texttt{random\_hue=[0.05]} \\
\bottomrule
\end{tabular}
\end{table*}

\begin{table*}
\centering
\caption{ACT hyperparameters for real-world experiments.}
\label{tab:act_hyperparameters_aloha}
\begin{tabular}{ll}
\toprule
hyperparameter & value \\
\midrule
\# GPUs & 1 x NVIDIA A100-SXM4-80GB GPU \\
learning rate (LR) & 1e-5 \\
total batch size & 8 \\
training strategy & full parameter tuning \\
input images & 1 third-person camera image, 1 wrist-mounted camera image \\
input image size & 224 x 224 px \\
use observation history & no (use single-step inputs) \\
action chunk size & 50 steps (predict 50, execute 25 open-loop at test time) \\
use proprio (robot state) & yes \\
image backbone & ResNet-18 \\
\# trainable parameters & 84M for ResNet-18 variant \\
\bottomrule
\end{tabular}
\end{table*}

\begin{table*}
\centering
\caption{OpenVLA hyperparameters for real-world experiments.}
\label{tab:openvla_oft_hyperparameters_aloha}
\begin{tabular}{ll}
\toprule
hyperparameter & value \\
\midrule
\# GPUs & 1 x NVIDIA A100-SXM4-80GB GPU \\
learning rate (LR) & 5e-4 \\
total batch size & 8 \\
training strategy & LoRA tuning, LoRA rank = 32 \\
input images & 1 third-person camera image, 1 wrist-mounted camera image \\
input image size & 224 x 224 px \\
use observation history & no (use single-step inputs) \\
action chunk size & 25 steps (predict 25, execute all 25 open-loop at test time) \\
use proprio (robot state) & yes \\
use FiLM & yes \\
\# trainable parameters & 853M total \\
image augmentations & 90\% random crops, color jitter: \\
& \;\; \texttt{random\_resized\_crop=dict(scale=[0.9, 0.9], ratio=[1.0, 1.0])} \\
& \;\; \texttt{random\_brightness=[0.2]} \\
& \;\; \texttt{random\_contrast=[0.8, 1.2]} \\
& \;\; \texttt{random\_saturation=[0.8, 1.2]} \\
& \;\; \texttt{random\_hue=[0.05]} \\
\bottomrule
\end{tabular}
\end{table*}

\begin{table*}
\centering
\caption{$\pi_{0.5}$ hyperparameters for real-world experiments.}
\label{tab:pi0_hyperparameters_aloha}
\begin{tabular}{ll}
\toprule
hyperparameter & value \\
\midrule
\# GPUs & 1 x NVIDIA A100-SXM4-80GB GPU \\
learning rate (LR) & 5e-5 peak LR (2K steps linear warmup)  \\
total batch size & 64 \\
training strategy & full parameter tuning \\
input images & 1 third-person camera image, 1 wrist-mounted camera image \\
input image size & 224 x 224 px \\
use observation history & no (use single-step inputs) \\
action chunk size & 10 steps (predict 10, execute all 10 open-loop at test time) \\
use proprio (robot state) & yes \\
\# trainable parameters & 3.3B \\
diffusion sampling algorithm & flow matching \\
number of integration steps & 10 \\
image augmentations & random crop (for non-wrist images), random rotation (for non-wrist images), color jitter: \\
& \;\; \texttt{augmax.RandomCrop(int(width * 0.95), int(height * 0.95))} \\
& \;\; \texttt{augmax.Rotate((-5, 5))} \\
& \;\; \texttt{augmax.ColorJitter(brightness=0.3, contrast=0.4, saturation=0.5)} \\
\bottomrule
\end{tabular}
\end{table*}

Following the backbone, VGGT utilizes multiple specialized decoding heads to map the latent tokens into a comprehensive set of geometric attributes. Specifically, the model jointly estimates four distinct categories of output for each frame $i$:
\begin{equation}
\mathcal{O}_i = \{g_i, D_i, P_i, T_i\}
\end{equation}
where $g_i$ denotes the 9-dimensional camera parameters, $D_i$ represents the dense depth map, $P_i$ signifies the viewpoint-invariant point map, and $T_i$ is the dense feature map for point tracking.

To infer the camera geometry, the input sequence is augmented with learnable camera tokens $t_{i,g}$. After these tokens evolve through the alternating-attention layers, the resulting representations $t_{i,g}^{(L)}$ are passed through an additional refinement block consisting of multiple self-attention layers. The final camera parameters are then regressed via a linear projection:
\begin{equation}
g_i = [q_i, t_i, f_i] \in \mathbb{R}^4 \times \mathbb{R}^3 \times \mathbb{R}^2
\end{equation}
where $q_i$ is the rotation quaternion, $t_i$ is the translation vector, and $f_i$ denotes the focal length. This head effectively decodes global geometric constraints into explicit extrinsic and intrinsic parameters.

To produce high-resolution geometric maps (e.g. dense depth map) from the sparse set of $K$ image tokens $t^{(L)}_{i,k}$, the model employs a Dense Prediction Transformer (DPT) module.
Specifically, intermediate tokens $\{t_i^{(s)}\}$ from multiple backbone stages are reassembled into spatial grids via a mapping $\phi_s$, and subsequently merged through a hierarchical fusion operator $\Psi$ to construct a dense feature map $F_i$. The final geometric attributes for each pixel $u$, including depth and point maps, are then derived through a linear projection:
\begin{equation}
V_i(u) = \mathbf{W} \left[ \Psi \left( \{ \phi_s ( t_i^{(s)} ) \}_{s \in \mathcal{S}} \right) \right]_u + \mathbf{b}
\end{equation}
where $V_i(u) = [D_i(u), P_i(u), \Sigma_i(u)]^\top$ denotes the vector of estimated 3D properties at coordinate $u$.

For point tracking, the model generates dense tracking features $T_i$ by further processing the image feature maps. For a specific query point $y_q$ in a source frame $I_q$, its corresponding location $\hat{y}_i$ in any target frame $I_i$ is determined by calculating the normalized correlation between the query feature and the target feature map:
\begin{equation}
\hat{y}_i = \text{argmax}_{y \in I_i} \left( \frac{\exp(T_q(y_q) \cdot T_i(y))}{\sum_{y' \in I_i} \exp(T_q(y_q) \cdot T_i(y'))} \right).
\end{equation}
In this formulation, the term within the argmax represents a dense similarity-based attention map, where the track position is localized by identifying the pixel that yields the maximum feature response across the target spatial domain.

\section{Theoretical Comparison: VGGT vs. VLM}

The theoretical Divergence between Visual Geometry Grounded Transformers (VGGT) and standard Vision-Language Models (VLM) begins with their divergent optimization manifolds. While both architectures utilize the attention mechanism, a VLM is typically optimized over a semantic manifold derived from a causal linguistic objective. Given a sequence of multimodal tokens, the VLM minimizes the negative log-likelihood of the next token prediction:
\begin{equation}
\mathcal{L}_{\text{VLM}} = -\sum_{i=1}^{N} \mathbb{E}_{(I, Y) \sim \mathcal{D}_{vlm}} \left[ \log P(y_i \mid y_{<i}, \Phi_{\text{enc}}(I)) \right]
\end{equation}
In this formulation, the visual observation $I$ is projected into a latent space where the distance metric represents conceptual or taxonomic similarity rather than physical constraints. Conversely, the VGGT framework is pre-trained to ground visual features within the metric properties of 3D Euclidean space. The objective functions for VGGT are designed to minimize the discrepancy between the latent representation and the true geometric structure $\mathcal{S}$ of the scene, often expressed as a reconstruction or consistency loss:
\begin{equation}
\mathcal{L}_{\text{VGGT}} = \int_{\Omega} \left\| \Psi(f_{\theta}(I), u, v) - \mathbf{X}_{3D}(u, v) \right\|^2 d\Omega
\end{equation}
where $\mathbf{X}_{3D}(u, v)$ denotes the ground-truth 3D coordinates at pixel $(u, v)$ and $\Psi$ is a differential geometric projection.

The architectural distinction is further refined through the Alternating Attention mechanism in VGGT, which contrasts with the uniform self-attention or cross-attention layers of the VLM. In VGGT, the spatial-temporal reasoning is decoupled across successive Transformer layers to preserve local geometric integrity while capturing global context. For a set of multi-view or multi-frame features $\{z_1, \dots, z_T\}$, the attention operations alternate as follows.
For even layers $2l$:
\begin{equation}
\hat{z}_t^{(2l)} = \text{Attn}(Q=z_t^{(2l-1)}, K=z_t^{(2l-1)}, V=z_t^{(2l-1)}), \forall t \in \{1, \dots, T\}
\end{equation}
For odd layers $2l+1$:
\begin{equation}
\hat{Z}^{(2l+1)} = \text{Attn}(Q=Z^{(2l)}, K=Z^{(2l)}, V=Z^{(2l)})
\end{equation}
In the local phase, the computation is restricted to the intra-frame domain, ensuring that the spatial features of each frame are refined independently. In the global phase, the sequence $Z = [z_1, \dots, z_T]$ is processed as a unified entity, allowing for inter-frame geometric alignment. This is structurally distinct from the VLM’s autoregressive attention, which is constrained by a triangular causal mask $M$ where $M_{ij} = 0$ if $i \ge j$ and $M_{ij} = -\infty$ otherwise, a restriction that is often counterproductive for capturing the non-directional nature of 3D spatial geometry.

From a control theory perspective, the modeling objective of VGGT is more naturally aligned with the generation of 3D actions in the robot's workspace. A robotic action $a$ resides in the Special Euclidean Group $\mathbb{SE}(3)$, and the learning task is to find a mapping $\mathcal{F}: \mathcal{M} \to \mathbb{SE}(3)$. We can characterize the efficiency of this mapping by examining the Lipschitz continuity of the transformation. For the VGGT latent manifold $\mathcal{M}_{\text{geo}}$, which is pre-aligned with 3D metric space, the Lipschitz constant $L$ is minimized:
\begin{equation}
\|\mathcal{F}(z_a) - \mathcal{F}(z_b)\|_{\mathbb{SE}(3)} \le L_{\text{VGGT}} \|z_a - z_b\|_{\mathcal{M}_{\text{geo}}}.
\end{equation}
Because $\mathcal{M}_{\text{geo}}$ preserves the equivariance properties of the physical world, the mapping to the action space is a low-complexity transformation. In contrast, the VLM's semantic manifold $\mathcal{M}_{\text{sem}}$ necessitates a significantly higher Lipschitz constant $L_{\text{VLM}} \gg L_{\text{VGGT}}$ to map abstract tokens to precise coordinate-based motor commands, leading to a more volatile optimization landscape.

Consequently, VGGT serves as a superior backbone for robotic manipulation for three primary reasons. First, its 3D-grounded objective ensures that the internal representations are inherently aware of the metric distances and volumes required for precise action prediction. Second, the Alternating Attention mechanism provides a specialized structural prior for fusing multi-view visual inputs and linguistic instructions without losing local spatial resolution. Third, the non-autoregressive nature of VGGT facilitates a parallelized inference process. This potentially allows for higher-frequency control loops. These factors collectively position VGGT as a more robust and efficient architecture for grounding linguistic commands into physical 3D interactions.

\section{Detailed Simulation Setup}

In this section, we provide a detailed description of the simulation setup and evaluation benchmark used in our experiments. All experiments are conducted on the LIBERO benchmark, a large-scale simulation platform designed for evaluating embodied agents on diverse household manipulation tasks.

\begin{figure}[t]
    \centering
    \includegraphics[width=0.98\linewidth]{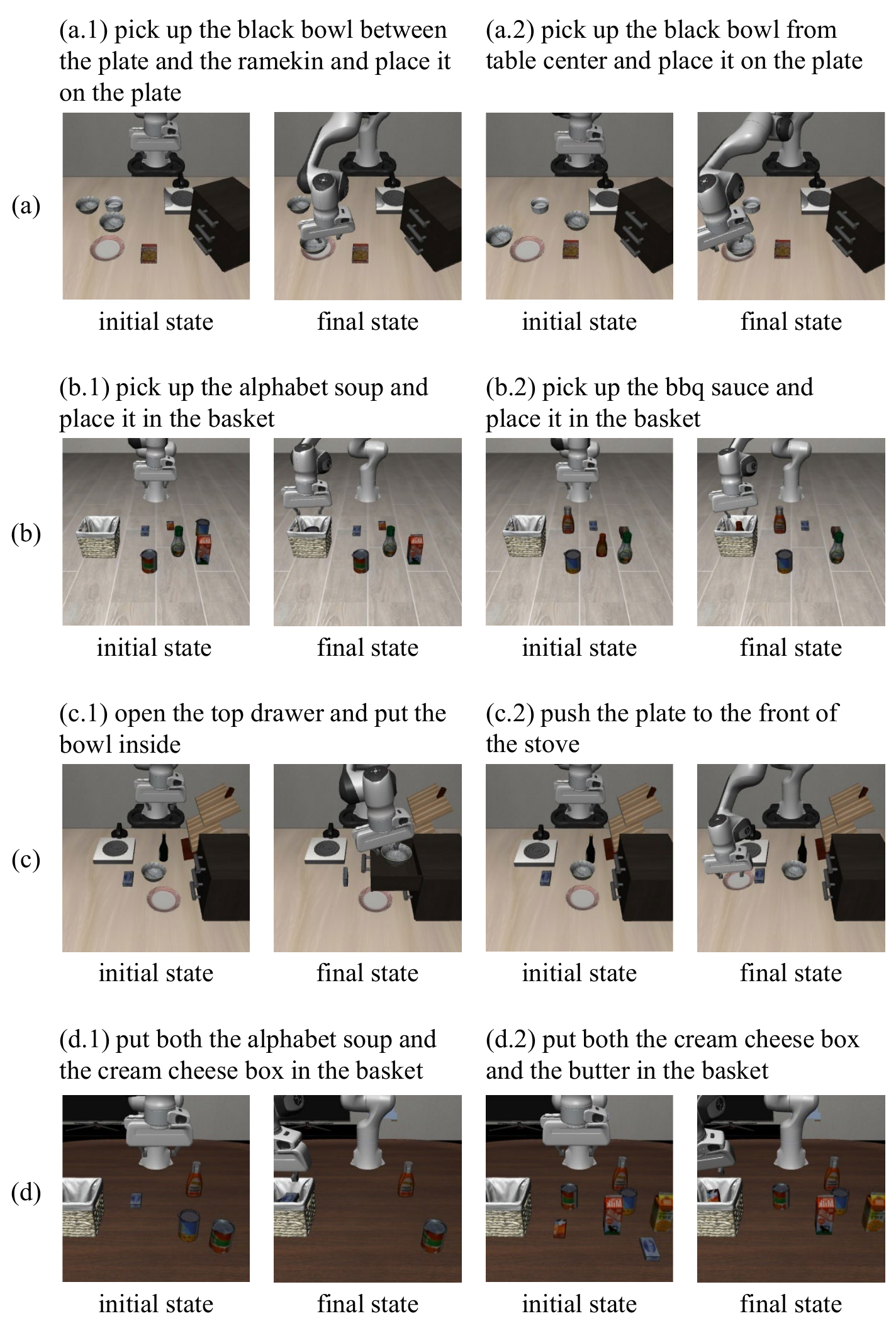}
    \vspace{-0.3cm}
    \caption{\underline{Representative tasks} from the four LIBERO suites: (a) LIBERO-Spatial, which focuses on spatial relationships, (b) LIBERO-Object, which evaluates object generalization, (c) LIBERO-Goal, which tests compositional task understanding, and (d) LIBERO-Long, which requires long-horizon sequential reasoning.}
    \vspace{-0.3cm}
    \label{fig:libero_tasks}
\end{figure}

Rather than forming a strict hierarchy of difficulty, the LIBERO benchmark organizes tasks into multiple suites, each targeting a distinct dimension of generalization. In this work, we evaluate our method on four representative suites: \textbf{LIBERO-Spatial}, \textbf{LIBERO-Object}, \textbf{LIBERO-Goal}, and \textbf{LIBERO-Long (LIBERO-10)}. These suites collectively assess the agent’s capabilities in spatial reasoning, object generalization, goal composition, and long-horizon planning. Representative tasks are illustrated in \cref{fig:libero_tasks}.

\subsection{Simulation Environment Setup}

The simulation environment consists of a tabletop manipulation setting with a robotic arm equipped with a parallel gripper. The scene contains both rigid objects (e.g., bowls, mugs, plates) and articulated structures (e.g., drawers, cabinets, and kitchen appliances).

The agent receives visual observations from multiple viewpoints, including a wrist-mounted camera and static cameras. In some configurations, depth observations are also provided. To improve robustness and prevent overfitting, object poses are randomly initialized within predefined regions, and small variations are introduced in camera viewpoints and lighting conditions.

The action space is continuous, consisting of end-effector position, orientation, and gripper open/close state. Each task is executed within a fixed time horizon, and success is determined based on task-specific completion criteria.

\begin{figure*}[t]
    \centering
    \includegraphics[width=0.9\linewidth]{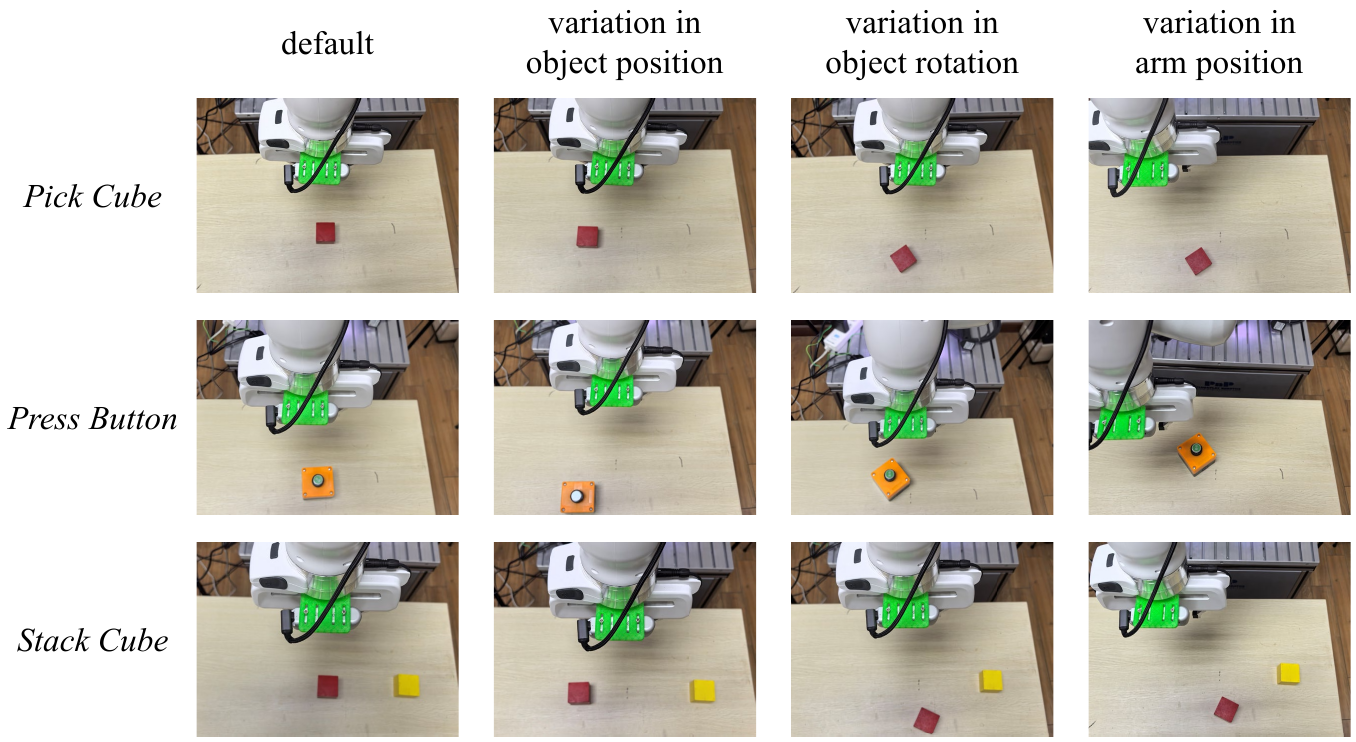}
    \vspace{-0.3cm}
    \caption{\underline{Initial state variations for evaluation tasks.} This figure illustrates the three evaluation tasks under four types of initial condition variations. Despite these diverse and challenging variations, our VGA consistently achieves high manipulation success rates across all settings.}
    \vspace{-0.3cm}
    \label{fig:task_variation}
\end{figure*}

\subsection{Task Suites}

\textbf{\emph{LIBERO-Spatial} Suite.}
The LIBERO-Spatial suite evaluates the agent’s ability to understand and execute spatial relationships between objects (see \cref{fig:libero_tasks}-(a)). Tasks typically involve relative placement instructions such as placing an object \emph{on}, \emph{inside}, or \emph{next to} another object. 
To succeed, the agent must accurately perceive relative positions and perform precise placement actions. These tasks emphasize fine-grained geometric reasoning and control. The object categories are generally fixed, while spatial configurations vary across episodes.

\noindent\textbf{\emph{LIBERO-Object} Suite.}
The LIBERO-Object suite focuses on generalization across different object instances (see \cref{fig:libero_tasks}(b)). While the task structure remains similar, the object categories may change significantly between training and evaluation.
This suite tests whether the agent can transfer manipulation skills across visually and geometrically diverse objects. Instead of relying on memorized appearances, the agent must leverage semantic and geometric priors to handle unseen objects.

\noindent\textbf{\emph{LIBERO-Goal} Suite.}
The LIBERO-Goal suite evaluates compositional generalization over task objectives (see \cref{fig:libero_tasks}(c)). Instead of varying object categories, this suite changes the combination of sub-goals and task requirements.
Tasks often involve multiple objects and constraints, requiring the agent to interpret complex instructions and decompose them into executable steps. Compared to the previous suites, this setting places greater emphasis on high-level reasoning and planning.

\noindent\textbf{\emph{LIBERO-Long} Suite.}
The LIBERO-Long suite, also known as LIBERO-10, consists of tasks that require long sequences of actions and complex temporal dependencies (see \cref{fig:libero_tasks}(d)). These tasks typically involve multiple stages, including interacting with articulated objects, moving objects across the scene, and satisfying sequential constraints.
The primary challenge lies in maintaining consistency over long horizons and avoiding error accumulation. The agent must plan over extended time steps and correctly sequence actions to achieve the final goal.

\section{Detailed Real-world Setup}

\subsection{Data Collection}

The robot setup features a 7-DoF Franka Panda arm equipped with a standard parallel-jaw gripper. The system operates within an 8-dimensional configuration and action space (7 joint positions + 1 gripper state). To capture comprehensive visual information, we employ two fixed camera views: a 3rd-person static camera providing a global view of the workspace, and a wrist-mounted camera providing egocentric observations. The physical configuration is illustrated in~\cref{fig:realworld_config}.

To facilitate high-fidelity data acquisition, we utilized the GELLO teleoperation framework~\cite{wu2024gello}, which leverages a 3D-printed master device for intuitive, joint-to-joint mapping of human demonstrations. This setup allowed us to efficiently collect 80 to 100 real-world trajectories per task. During the demonstration phase, we systematically introduced stochasticity by varying the initial robotic arm poses and object placements within a localized spatial range. This variation ensures the model learns to adapt to diverse starting configurations rather than memorizing a single static path, thereby improving the robustness of the resulting policy.

Data quality was maintained through a rigorous filtering process aimed at optimizing the training signal. We exclusively retained trajectories that resulted in a successful task completion, immediately discarding any failed attempts or collisions. To ensure the demonstrations were smooth and efficient, we further manually removed outliers characterized by unstable or jerky movements. Additionally, we filtered out trajectories with abnormal durations—either excessively long or unnaturally short—to guarantee temporal consistency. This pruning resulted in a refined dataset of high-quality, goal-oriented demonstrations suitable for training the VGGT backbone in complex manipulation scenarios.

\subsection{Evaluation Tasks}

In this section, we provide additional details on the real-world evaluation tasks and their experimental setups. All tasks are executed within a fixed horizon of 800 control steps, and a trial is considered a failure if the success condition is not met within this limit.

The \emph{Pick Cube} task requires the robot to grasp a cube from the tabletop and lift it to a predefined height. Success is achieved when the cube is stably grasped and raised above a vertical threshold. The \emph{Press Button} task requires the robot to reach a designated mechanical button and apply force to activate it; success is defined by a clear state change of the button. The \emph{Stack Cube} task requires placing one cube on top of another; success is achieved when the cube is placed with sufficient alignment and remains stable without collapsing.
Each trial is allowed to proceed for up to 1200 control steps (approximately 2 minutes), and failure is declared if the success condition is not met within this limit.

To systematically evaluate robustness under diverse initial conditions, we introduce controlled variations in the task setup along multiple axes. Specifically, for each task, we generate different initial configurations by randomizing the object position on the tabletop within a predefined workspace, perturbing the initial orientation of the objects, and altering the initial pose of the robotic manipulator. These variations are designed to induce a broad distribution of starting states while remaining within feasible operational bounds of the system. The accompanying visualization (a $3 \times 4$ grid of~\cref{fig:task_variation}) illustrates these variations for each task: the first column shows the default setup, while the remaining columns correspond to randomized object positions, randomized object orientations, and perturbed initial robot poses, respectively. This design ensures that the evaluation is not limited to a narrow set of canonical configurations, but instead reflects a more realistic and challenging distribution of manipulation scenarios.

\section{Limitations}
While our VGA model excels at precise spatial manipulation and cross-view generalization, it has limitations in tasks requiring commonsense knowledge or language reasoning.
This limitation arises from the fact that its backbone is not based on a large language model.
For example, instructions such as ``place the cube on the picture of Taylor Swift'' (see OpenVLA~\cite{kim2024openvla}) are challenging for VGA.
Incorporating external reasoning modules is a potential future direction to address this limitation.
\clearpage


\end{document}